\documentclass{article}

\PassOptionsToPackage{round}{natbib}

\usepackage[preprint]{neurips_2022}

\usepackage[utf8]{inputenc} 
\usepackage[T1]{fontenc}    
\usepackage{hyperref}       
\usepackage{url}            
\usepackage{booktabs}       
\usepackage{amsfonts}       
\usepackage{nicefrac}       
\usepackage{microtype}      
\usepackage{xcolor}         
\usepackage{xspace}
\usepackage{amsmath}
\usepackage{amssymb}
\usepackage{pifont}
\usepackage{mathtools}
\usepackage{amsthm}
\usepackage{bm}
\usepackage{wrapfig}
\usepackage{multirow}
\usepackage[normalem]{ulem}
\usepackage{cleveref}
\usepackage{graphicx}
\usepackage{caption, subcaption}
\usepackage{paralist}
\usepackage{ulem}
\usepackage[english]{babel}

\usepackage{etoolbox}
\makeatletter
\patchcmd{\hyper@makecurrent}{%
    \ifx\Hy@param\Hy@chapterstring
        \let\Hy@param\Hy@chapapp
    \fi
}{%
    \iftoggle{inappendix}{
        \@checkappendixparam{chapter}%
        \@checkappendixparam{section}%
        \@checkappendixparam{subsection}%
        \@checkappendixparam{subsubsection}%
        \@checkappendixparam{paragraph}%
        \@checkappendixparam{subparagraph}%
    }{}%
}{}{\errmessage{failed to patch}}

\newcommand*{\@checkappendixparam}[1]{%
    \def\@checkappendixparamtmp{#1}%
    \ifx\Hy@param\@checkappendixparamtmp
        \let\Hy@param\Hy@appendixstring
    \fi
}
\makeatletter

\newtoggle{inappendix}
\togglefalse{inappendix}

\apptocmd{\appendix}{\toggletrue{inappendix}}{}{\errmessage{failed to patch}}

\newcommand\myshade{50}
\colorlet{mylinkcolor}{blue}
\colorlet{mycitecolor}{green}
\colorlet{myurlcolor}{red}
\hypersetup{
  linkcolor  = mylinkcolor!\myshade!black,
  citecolor  = mycitecolor!\myshade!black,
  urlcolor   = myurlcolor!\myshade!black,
  colorlinks = true,
}

\usepackage{amsmath,amsfonts,bm}

\def\eqref#1{equation~\ref{#1}}

\def\1{\bm{1}}

\DeclareMathAlphabet{\mathsfit}{\encodingdefault}{\sfdefault}{m}{sl}
\SetMathAlphabet{\mathsfit}{bold}{\encodingdefault}{\sfdefault}{bx}{n}

\newcommand{\knn}{$k$NN\xspace}

\newcommand{\knnlm}{\knn{}-LM\xspace}
\newcommand{\knnlms}{\knnlm{}s\xspace}

\usepackage{array}
\newcolumntype{H}{>{\setbox0=\hbox\bgroup}c<{\egroup}@{}}

\title{Why do Nearest Neighbor Language Models Work?}

\author{%
  Frank F. Xu\quad Uri Alon\quad  Graham Neubig \\
  Language Technologies Institute\\
  Carnegie Mellon University \\
  \texttt{\{fangzhex,ualon,gneubig\}@cs.cmu.edu}
}

\begin{document}

\maketitle

\begin{abstract}
Language models (LMs) compute the probability of a text by sequentially computing a representation of an already-seen context and using this representation to predict the next word.
Currently, most LMs calculate these representations through a neural network consuming the immediate previous context.
However recently, \emph{retrieval-augmented LMs} have shown to improve over standard neural LMs, by accessing information retrieved from a large datastore, in addition to their standard, parametric, next-word prediction.
In this paper, we set out to understand \emph{why} retrieval-augmented language models, and specifically why $k$-nearest neighbor language models (\knnlms) perform better than standard parametric LMs, even when the $k$-nearest neighbor component retrieves examples from the same training set that the LM was originally trained on.
To this end, we perform a careful analysis of the various dimensions over which  \knnlm~diverges from standard LMs, and investigate these dimensions one by one.
Empirically, we identify three main reasons why \knnlm performs better than standard LMs: using a different input representation for predicting the next tokens, \emph{approximate} \knn search, and the importance of softmax temperature for the \knn distribution.
Further, we incorporate these insights into the model architecture or the training procedure of the standard parametric LM, improving its results without the need for an explicit retrieval component.
The code is available at \url{https://github.com/frankxu2004/knnlm-why}.
\end{abstract}

\section{Introduction}
Language modeling is the task of predicting the probability of a text (often conditioned on context), with broad-spanning applications across natural language processing~\citep{bengio2003neural,merity2018regularizing,baevski2018adaptive,brown2020language}.
This modeling is usually done by sequentially encoding a context $c_t$ using a trained neural network function $f$, and computing the probability of the next word $w_t$ according to $f\left(c_t\right)$ and a vector representation of $w_t$.

Recently, \emph{retrieval-augmented} LMs have shown a series of impressive results~\citep{grave2017unbounded,guu2018generating,he2020learning,khandelwal20generalization,borgeaud2022improving,alon2022neuro}. 
Retrieval-augmented LMs compute next token distributions based not only on the immediately preceding context $c_t$ and the model parameters, 
but also on an external datastore, from which examples are retrieved and incorporated into the base LM's prediction.

One retrieval-augmented model that is notable for both its simplicity and efficacy is the $k$-nearest neighbor language model \citep[\knnlm;][]{khandelwal20generalization}.
It extends a trained base LM by linearly interpolating the output word distribution with a \knn model. 
The nearest neighbors are retrieved according to the distances between the current context embedding of the base LM and all the context embeddings in the datastore. 
The datastore is created by encoding all contexts from any text collection, including the original LM training data.

\begin{figure}[t]
\centering
\includegraphics[width=0.7\textwidth]{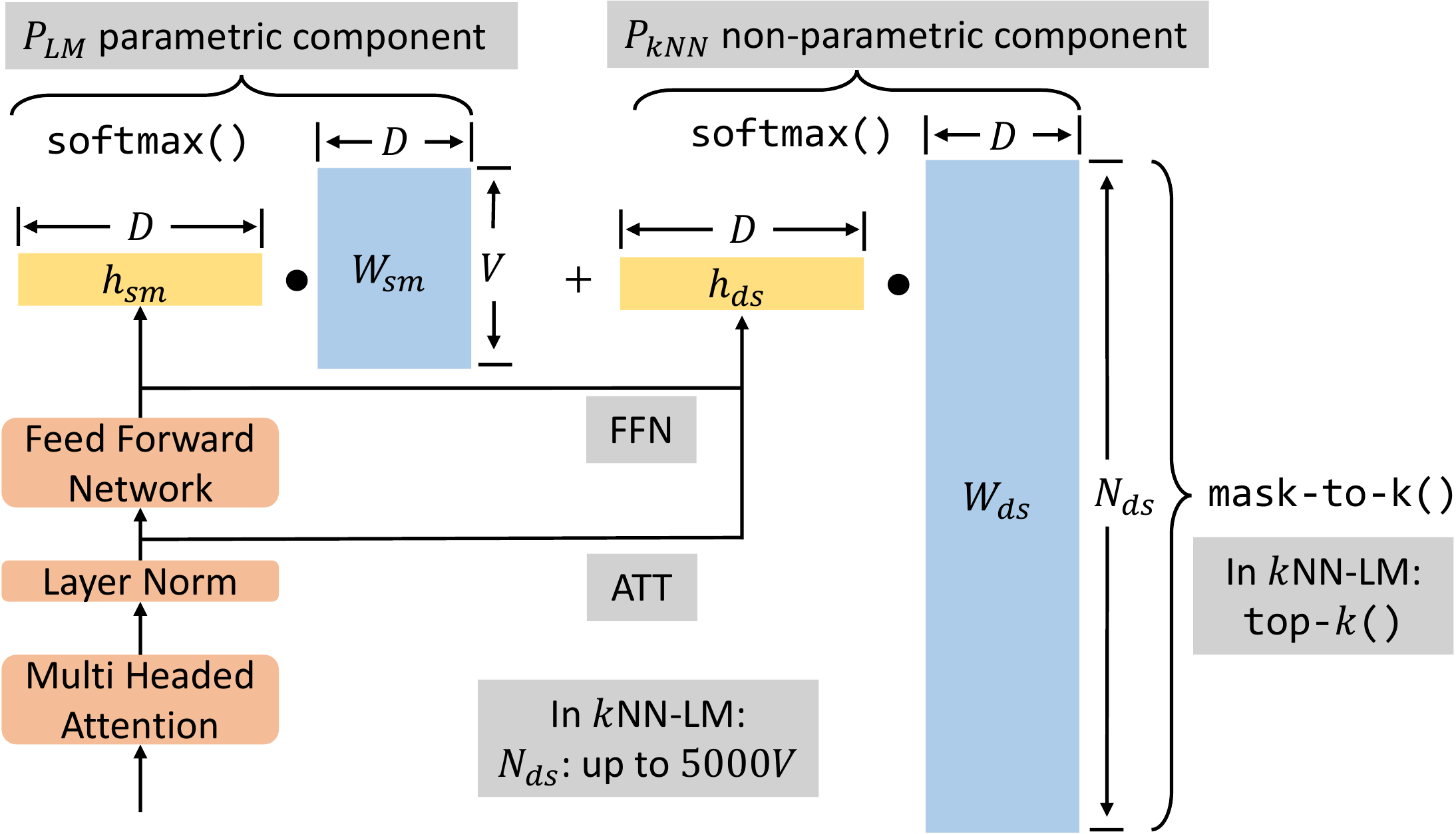}
\caption{An illustration of the generalized formulation of \knnlm in \autoref{eqn:general}.}
\label{fig:overview}
\end{figure}

One of the most surprising results from \citet{khandelwal20generalization} is that \knnlm reduces the perplexity of the base LM \emph{even when the \knn component is retrieving examples from the same training set that the LM was originally trained on}, indicating that the \knnlm improves the ability to model the training data and is not simply benefiting from access to more data.
Intrigued by this, we ask questions like, could \knnlm be improving because of capacity issues in the parametric base LM?
In this paper, we set out to understand why \knnlms work even in this setting.

In the following sections,
we first elucidate connections between the added \knn component and the standard LM component.
Specifically, we note that word distributions from the two components are both calculated using a softmax function, based on the similarity of the current context embedding with a set of embeddings that corresponds to different next words.
With this intuition, we formalize and generalize the non-parametric distribution calculation with the softmax layer and word embedding layer used in parametric LMs.
We then show that this generalized form exposes a variety of design choices, e.g., the number of context embeddings in the datastore, the input representation used in softmax layer, different similarity functions, as well as the approximation and sparsification implementations in the \knn search.  
This provides a general framework for analyzing \knnlm and similar models and allows us to perform ablation studies that test the importance of various design decisions.

We proceed to propose multiple hypotheses for why \knnlm works, which are testable by adjusting the various parameters exposed by our generalized formulation.
Based on these hypotheses, we perform ablation experiments and analyze the nuances between different implementations of the generalized version of $P_{kNN}$.
As the answer to our question, ``why \knnlms work'', we eventually show that the most probable reasons are threefold:
\begin{enumerate}
    \item Ensembling the output of softmax using two  representations from different layers of the transformer is important; in our experiments, this accounts for 55\% of the performance gain of \knnlm, or 6.5\% relative perplexity improvement compared to the base LM.
    \item 
    \knnlm uses \emph{approximate} nearest neighbor search to handle the large number of candidates, and the lack of this preciseness in this algorithm actually helps \knnlm to generalize \emph{better} than using exact nearest neighbor search and distance calculation, possibly due to a regularization effect.
    The relative perplexity improvement from this factor is about 2.6\%.
    \item Depending on the design decisions that are chosen for modeling, adding a temperature term to the \knn non-parametric component can become crucial to the success of modeling (although coincidentally, in the original settings of \citet{khandelwal20generalization}, a temperature of 1.0 is close to optimal, which hid the importance of this term).
    In some settings, the relative perplexity gap between the default and optimal temperature can be as high as 3.7\%.
\end{enumerate}

Finally, one significant drawback to the current \knnlm is the inefficiency of \knn search performed at each step  \citep{he2021efficient,borgeaud2022improving,alon2022neuro,Wang2022EfficientCK}.
Because of the similarity between \knnlm and the parametric LM's last layers and the many design choices, we also demonstrate that we are able to make \knnlm more efficient by substituting the \knn search with another matrix operation that can fit in accelerator memory while maintaining more than half the perplexity improvement, or more than 6.5\% relative improvement compared to the base LM.



\section{Formalizing and Generalizing \knnlm}
\label{sec:alternative}
\knnlm~\citep{khandelwal20generalization} is a linear interpolation between a base LM and a \knn model.
Given a set of contexts $c_i$ and their corresponding next token $w_i$ as a pair $(c_i, w_i) \in \mathcal{D}$, \knnlms create a datastore $(\mathcal{K}, \mathcal{V}) = \{(k_{i}, v_{i})\}$, as a set of key-value pairs:
\begin{equation}
(\mathcal{K}, \mathcal{V})=\left\{\left(f\left(c_{i}\right), w_{i}\right) \mid\left(c_{i}, w_{i}\right) \in \mathcal{D}\right\}
\end{equation}

During inference, the parametric component of the LM generates the output distribution $p_{LM}(w_t|c_t;\theta)$ over the next tokens and produces the corresponding context representation $f(c_t)$, given the test input context $c_t$. 
Then the non-parametric component of the LM queries the datastore with the $f(c_t)$ representation
to retrieve its $k$-nearest neighbors $\mathcal{N}$ according to a distance function $d(\cdot,\cdot)$.
Next, the \knnlm computes a probability distribution over these neighbors using the softmax of their negative distances, and 
aggregates the probability mass for each vocabulary item across all of its occurrences in the retrieved targets:
\begin{align}
p_{\mathrm{kNN}}(w_t|c_t) &\propto \sum_{(k_{i}, v_{i}) \in \mathcal{N}} \textbf{1}_{w_t=v_{i}} \exp (-d(k_{i}, f(c_t)))
\label{eqn:pknn}
\end{align}
Finally, this distribution is interpolated with
the parametric LM distribution $p_{\mathrm{LM}}$ to produce the final \knnlm distribution:
\begin{align}
p(w_t|c_t;\theta) &= (1-\lambda) p_{\mathrm{LM}}(w_t|c_t;\theta) + \lambda p_{\mathrm{kNN}}(w_t|c_t)
\end{align}
where $\lambda$ is a scalar that controls the weights of the interpolation between two components, with higher $\lambda$ putting more weight on the non-parametric component.

Looking closely at \autoref{eqn:pknn}, we can notice a similarity between the calculation of $P_{kNN}$ and the standard $P_{LM}$.
The \knn distribution is based on the distances between the current context and the nearest neighbors from the datastore, normalized by a softmax function.
Recall that in (standard) parametric language models, the distribution over the vocabulary is also based on a measure of distance, the inner product between the current context embedding and the word embeddings of every token in the vocabulary.
Because each context embedding in the datastore $\left(\mathcal{K}, \mathcal{V}\right)$ corresponds to a target token, we can also view this datastore as a large word embedding matrix with multiple word embeddings for each of the vocabulary words.
Theoretically, given unlimited computation, we could calculate the distribution based on the distances to every embedding in the datastore, and aggregate by vocabulary items, making it more closely resemble $P_{LM}$.
In this case, $k=|\mathcal{D}|$, the size of the entire datastore, and~\autoref{eqn:pknn} becomes the following, based on the distances to every context in the datastore $\mathcal{D}$ instead of a subset of nearest neighbors $\mathcal{N}$.
\begin{align}
p_{\mathcal{D}}(w_t|c_t) &\propto \sum_{(k_{i}, v_{i}) \in \mathcal{D}} \textbf{1}_{w_t=v_{i}} \exp (-d(k_{i}, f(c_t)))
\label{eqn:unlimited_pknn}
\end{align}
In practice, we use \knn search as a way of approximation, by limiting the calculation to only $k$ nearest neighbors to avoid the computational cost of calculating the distribution over the entire datastore.

If we re-write and generalize \autoref{eqn:pknn}, both the \knnlm of \citet{khandelwal20generalization} and a large number of related models can be expressed through the following equation:
\begin{equation}
\label{eqn:general}
P_\text{interp} = (1-\lambda) \underbrace{\text{softmax}(W_{sm} \cdot h_{sm})}_{P_{LM}\text{ parametric component}} + \lambda \underbrace{M \text{softmax}( \text{mask-to-k}(W_{ds} \otimes h_{ds} ) / \tau)}_{P_{kNN}\text{ non-parametric component}}.
\end{equation}
\autoref{fig:overview} provides an illustration of \autoref{eqn:general}.
The first term of the equation is the standard parametric language model, whereas the second represents a generalized version of utilizing an external datastore.
The first component, the output layer of a common parametric language model, is relatively straightforward.
$W_{sm}$ of size $V \times D$ is the embedding matrix of the output token, and $h_{sm}$ is the context vector used to calculate the distribution of the output token, usually the output of the final feedforward layer in the transformer.

In the second component,
$W_{ds}$ represents the datastore, of size $N_{ds}  \times D$.
$N_{ds}$ is the number of entries in the datastore, and $D$ is the size of each context vector.
$h_{ds}$ represents the context vector used to query the datastore. 
As shown in~\autoref{fig:overview}, these vectors can come from different layers of the transformer architecture.
$\otimes$ represents the operation type used to calculate the similarity between context vectors and the query vector, which also has several alternatives that we discuss below.

$\text{mask-to-k}(\cdot)$ represents a function to sparsify similarity scores across the datastore, setting all but $k$ similarity scores to $-\infty$, which results in probabilities of zero for all masked similarity scores after the softmax.
Practically, this is necessary for \knnlm{}s because the size of the datastore $N_{ds}$ makes it infeasible to calculate all outputs at the same time.
With masked logits, we apply a more generalized version of softmax with temperature $\tau$.
Intuitively adding the temperature can adjust the peakiness or confidence of the softmax probability distribution output.
After the softmax, the matrix $M$ of dimension $V \times N_{ds}$ sums the probability of the $N_{ds}$ datastore entries corresponding to each of the $V$ vocabulary entries.
Each column in this matrix consists of a one-hot vector with a value of 1 and the index corresponding to the vocabulary item $w_i$ corresponding to the datastore entry for $c_i$.

Within this formulation, it becomes obvious that there are many design choices for \knnlm-like models.
One important thing to note is that the right side of \autoref{eqn:general} is actually very similar to the left side representing the standard parametric language model, with a few additional components: $M$, mask-to-k, and $\otimes$.
More specifically, some of the design decisions that go into the \knnlm, and parallels with standard parametric models are:
\begin{enumerate}
    \item \textbf{Size of $W_{ds}$:} In the standard parametric model, the size of $W_{sm}$ is $V$ embedding vectors, each with $D$ dimensions. In the \knnlm the size of $W_{ds}$ is very large: $N_{ds}$, the size of the datastore, usually the number of tokens in the entire training corpus.
    \item \textbf{Input representation:} In the parametric model, $h_{sm}$ is the output from the feedforward layer in the last transformer block, which we abbreviate ``ffn''. 
    In contrast, \citet{khandelwal20generalization} rather use as $h_{ds}$ the output from the multi-headed attention layer of the last transformer block (before running the representations through the feed-forward network, and \emph{after} the LayerNorm~\citep{ba2016layer}), which we abbreviate as ``att''.
    \item \textbf{Similarity \& Temperature:} In the parametric model, the functional form of $\otimes$ is the inner product (abbreviated IP), whereas \citet{khandelwal20generalization} use negative squared L2 distance (abbreviated L2) as a similarity function between $W_{ds}$ and $h_{ds}$. As the similarity scores are turned into probability distributions with the softmax function, the choice of softmax temperature ($\tau$) can control the scaling of the similarity scores and thus the peakiness of the non-parametric component distribution. 
    \item \textbf{Approximation \& Sparsification:} In the parametric model, $k=V$, and no values are masked, but in the \knnlm, $k \ll V$, and most of the datastore entries are pruned out. The definition of the $\text{mask-to-k}(\cdot)$ function, i.e. how to select the important datastore embeddings to include in the similarity calculation (in \knnlm's case the $k$ nearest neighbors), is a crucial open design choice.
\end{enumerate}

In the following sections, we set out to better understand how each of these design decisions contributes to the improvement in accuracy due to the use of \knnlms.

\section{Baseline \knnlm Results}
\label{sec:baseline}

First, we evaluate the \knnlm baseline on the Wikitext-103 dataset~\citep{merity2016pointer}, and examine the importance of two design choices: the input representation $h_{ds}$ and the similarity function $\otimes$.

In models examined in this paper, the parametric model is a transformer language model with mostly the same architecture as in~\cite{khandelwal20generalization}.
However, We do make modifications to the original base LM~\citep{baevski2018adaptive} to accommodate our experimentation need.
We using BPE tokenization~\citep{sennrich2015neural} to train a smaller vocabulary (33K) than the original (260K) on the training corpus of Wikitext-103, as subword tokenization is ubiquitous in many state-of-the-art language models~\citep{radford2019language,devlin2018bert,liu2019roberta,brown2020language}.
Using subword tokenization also eliminates the need for adaptive softmax~\citep{joulin2017efficient}.
It makes the output layer more generalized, sharing more resemblance to the \knn component as described in~\autoref{sec:alternative}, and facilitates the ablation studies in this paper.%
\footnote{By training our own version of the base LM from scratch with BPE tokenization and a standard output softmax layer, our LM's perplexity is worse than that used in the original \knnlm paper. However, we observe similar relative gains from the additional \knn component. We argue that the base LM's performance is orthogonal to the study of the factors behind \knnlm's improvements.}
This base LM has 268M parameters.
To get a perspective on how large the datastore is, it is built on the training data that contains nearly 150M BPE tokens, each paired with a context vector of size 1024. 
This datastore has a total memory consumption of about 300GB.
At every retrieval step, we take the top 1024 nearest neighbors, i.e., $k=1024$, following \citet{khandelwal20generalization}.
The interpolated perplexity is computed with optimal interpolation parameter $\lambda$ tuned according to the perplexity on the development set.
$\lambda$ is fixed during the inference for all predictions, the same as the standard \knnlm. 

\begin{table}[ht]
\small
\centering
\begin{tabular}{ccHcccHccHH}
\toprule
 & $h_{ds}$ & $N_{ds}$ & $\otimes$ & {}+\#params      & PPL      & $\lambda$ & Interp. PPL & Oracle & Overlap & Helping \\ 
 \midrule
Base LM         & -        & -        & -         & 0                & 21.750   & -         & -           & -      & -              & -             \\
\knnlm-L2        & att      & Big      & L2        & $N_{ds}\times D$ & $\infty$ & 0.271     & 19.174      & 14.230 & 1.000          & 0.380         \\
\knnlm-IP         & att      & Big      & IP        & $N_{ds}\times D$ & $\infty$ & 0.266     & 19.095      & 14.077 & 0.890          & 0.402         \\
\knnlm-L2         & ffn      & Big      & L2        & $N_{ds}\times D$ & $\infty$ & 0.065     & 20.734      & 15.594 & 0.600          & 0.357         \\
\knnlm-IP         & ffn      & Big      & IP        & $N_{ds}\times D$ & $\infty$ & 0.050     & 21.101      & 16.254 & 0.586          & 0.340         \\ 
\bottomrule
\end{tabular}
\caption{Performance of the parametric language model and several \knnlm variants.}
\label{tab:knn_baselines}
\end{table}

Results comparing multiple \knnlm variants are shown in \autoref{tab:knn_baselines}.
The first row represents the base parametric language model's perplexity.
The second is a formulation analogous to that of \citet{khandelwal20generalization}, and in the remaining rows, we vary the input representation $h_{ds}$ and distance function $\otimes$ from \autoref{eqn:general}.
All of them use a large datastore with size $N_{ds}$, approximately 5000 times the size of the vocabulary $V$, as also reflected in ``{}+\#params'', the number of \emph{additional} parameters other than the base LM.

We report several important quantities with respect to each model.
\begin{itemize}
\item ``PPL'' shows the perplexity of \emph{only} the \knn component of the model $p_{\mathrm{kNN}}()$. This is $\infty$ for all \knnlm models in all cases, as when the \knn search does not retrieve any datastore entries corresponding to the true target word $w_t$ the probability of the target word will be zero.
\item ``Oracle'' shows the lower bound of the interpolation performance by choosing the best $\lambda$ for each token in the evaluation dataset, which will either be $\lambda=0$ or $\lambda=1$ depending on whether $P_{LM}(w_t|c_t)>P_{knn}(w_t|c_t)$ or not, respectively.
\end{itemize}

From the table, we can see that:

\begin{enumerate}
    \item Using the output of the multi-headed attention layer (``att'') as $h_{ds}$ (instead of the standard ``ffn'' layer) is crucial for better performance of \knnlm.
    \item In general, using negative squared L2 distance or inner product as a similarity function does not result in a large and consistent difference, although in our setting, IP provides slightly better performance when using the ``att'' inputs, and slightly worse when using ``ffn'' inputs.
    \item Interestingly, when using ``ffn'' and ``IP'', the same input and distance metric used in the parametric model, the results are the worst, indicating that the \knnlm is particularly benefiting when the \knnlm achieves a \emph{different view} of the data from the parametric model.
\end{enumerate}

We found in preliminary experiments that \knnlm is generalizable to other base language models as well, ranging from small models with 82M parameters to larger models with 774M parameters.
The gain from \knnlm is more significant when used with a smaller, less capable base language model, as expected. The details are shown in~\autoref{app:generalization-knnlm}.  
In this paper, we are mainly focused on the factors contributing to the relative improvements from \knnlm, instead of the absolute performance, so we use the 268M model for the remainder of the paper.

In the next sections, we perform further experiments with ablations on the general formulation~\autoref{eqn:general} to elucidate the key elements contributing to the performance improvements in \knnlm.

\section{Effect of Different $W_{ds}$ Formulations}
\label{sec:input_representation}

\subsection{Replacing the Datastore with Trainable Embeddings}
From the observation in~\autoref{sec:baseline}, we see that the choice of $h_{ds}$ has a large impact on the performance of \knnlm.
This intrigues us to explore if one key to the improvements afforded by \knnlm lies in the use of different input representations together, namely the attention output ($h_{ds}=\text{att}$) and feedforward output ($h_{ds}=\text{ffn}$).
However, from only the experiments above, it is not possible to disentangle the effect of the choice of $h_{ds}$ and that of other design choices and factors in \autoref{eqn:general}.

To test the effect of $h_{ds}$ in a more controlled setting, we remove the non-parametric datastore entirely, and initialize $W_{ds}$ in~\autoref{eqn:general} with a randomly initialized word embedding matrix with the same size ($N_{ds}=V$) as the LM's output embedding $W_{sm}$, and train $W_{ds}$ with all other parameters fixed.%
\footnote{
Because we previously found little difference between IP and L2 as similarity functions, we use IP in the experiments. 
For simplicity, we set temperature $\tau=1$.
}
The loss function for training is the cross-entropy loss of $\text{softmax}(W_{ds} \cdot h_{ds} )$ with respect to the ground-truth tokens, identically to how the base LM is trained.
We compare how using $h_{ds}=\text{att}$ or $h_{ds}=\text{ffn}$ affects the interpolated performance.
The results are shown in~\autoref{tab:input_representation}, and we also show results from \knnlms using these two varieties of input representation for reference.

From these experiments we can find several interesting conclusions:

\textbf{Effectiveness of re-training $W_{ds}$:} In the case of ``Learned $W_{ds}$ w/ FFN'', we are essentially re-learning the weights feeding into the softmax function separately from the underlying LM encoder. Despite this fact, we can see the model achieves a PPL of 20.920, which is 0.83 points better than the base model. This suggests that there is some benefit in learning the parameters of $W_{ds}$ after freezing the body of the transformer encoder.

\textbf{Effectiveness of ensembling two predictors:} In both cases for $W_{ds}$, the interpolated perplexity is significantly better than that of using a single predictor.
This is particularly the case when using the ``att'' representation for $h_{ds}$, suggesting that the utility of ensembling predictions from two views of the data is not only useful when using \knnlm, but also in standard parametric models as well.

\textbf{Parametric ensembles as an alternative to \knnlm?:}
Overall, by using a separate word embedding matrix with size $V\times D$ as an alternative to \knn, we can recover about 55\% of the performance gain achieved by \knnlm, with only a limited number of parameters and without the necessity for slow \knn retrieval every time a token is predicted.
This suggests that the majority of the gain afforded by \knnlm could be achieved by other more efficient means as well.


\begin{table}[h]
\centering
\small
\begin{tabular}{ccccccHccHH}
\toprule
 & $h_{ds}$ & $N_{ds}$ & $\otimes$ & {}+\#params      & PPL      & $\lambda$ & Interp. & Oracle & Overlap & Helping \\ \midrule
Base LM         & -        & -        & -         & 0                & 21.750   & -         & -           & -      & -       & -       \\
\knnlm w/ ATT         & att      & Big      & IP        & $N_{ds}\times D$ & $\infty$ & 0.266     & 19.095      & 14.077 & 0.890   & 0.402   \\
Learned $W_{ds}$ w/ ATT      & att      & 1x       & IP        & $V\times D$      & 22.584   & 0.407     & 20.353      & 16.954 & 0.541   & 0.483   \\
\knnlm  w/ FFN         & ffn      & Big      & IP        & $N_{ds}\times D$ & $\infty$ & 0.050     & 21.101      & 16.254 & 0.586   & 0.340   \\
Learned $W_{ds}$ w/ FFN        & ffn      & 1x       & IP        & $V\times D$      & 20.920   & 0.607     & 20.694      & 18.772 & 0.536   & 0.432   \\
\bottomrule
\end{tabular}
\caption{Performance comparison how the choice of $h_{ds}$, input representation, affects \knn baselines and models with learnable embeddings as datastore alternative. $h_{ds}$ is the attention layer output.}
\label{tab:input_representation}
\end{table}

\subsection{Increasing the Softmax Capacity}
\label{sec:bottleneck}

\begin{figure}[th]
\centering
\includegraphics[width=0.5\textwidth]{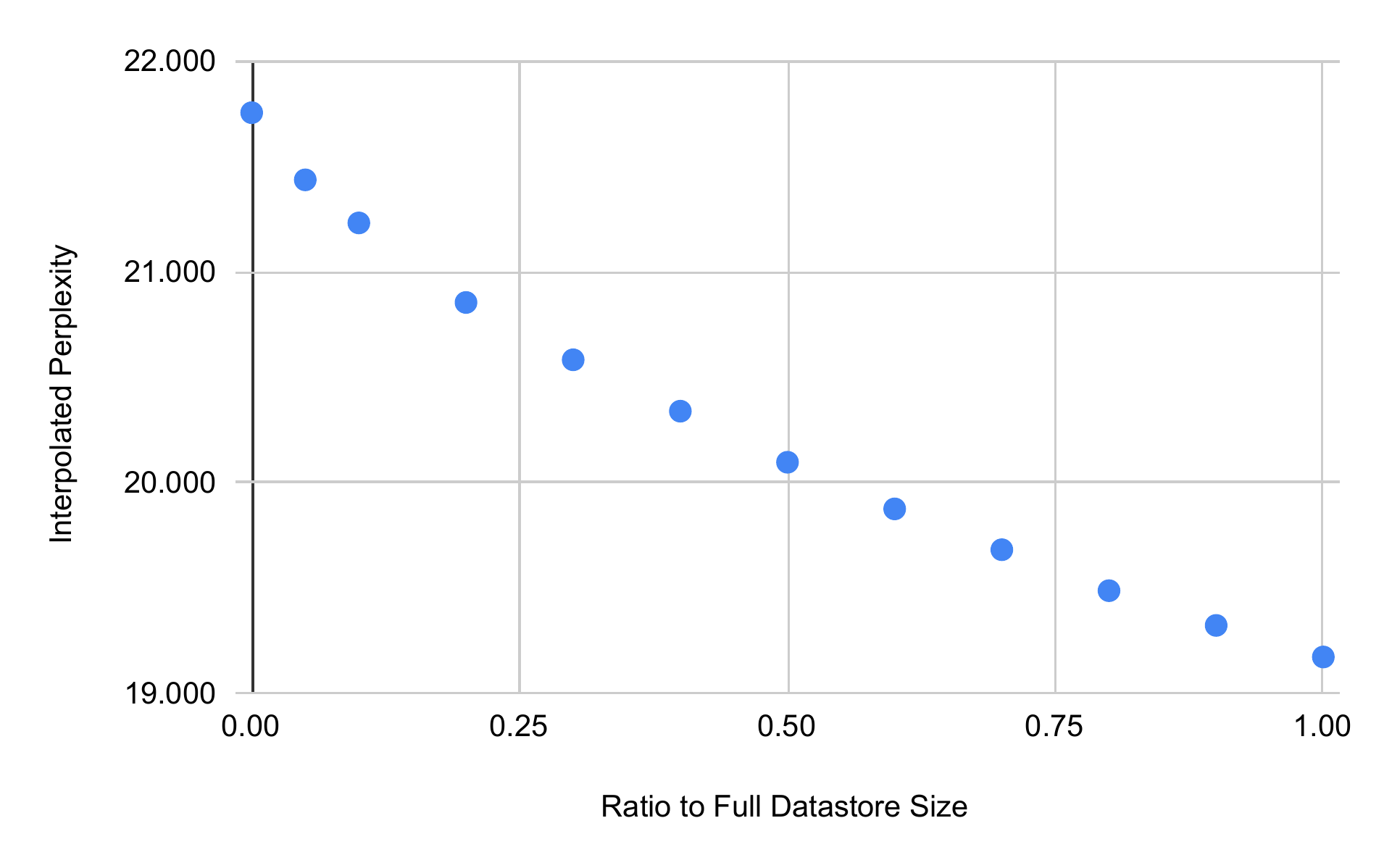}
\caption{The effect of the size of the datastore used for \knn retrieval on final interpolated perplexity.}
\label{fig:subsample_dstore_only}
\end{figure}

One premise behind \knnlm is that the large datastore is the key reason for the model working well: the larger the softmax capacity, the better the performance. 
Naturally, as a first step, we need to check whether such a big datastore is warranted and whether the high rank of $W_{ds}$ leads to better performance. 
We test the effect of the datastore size for \knn retrieval on \knnlm interpolated perplexity.
If a bigger datastore (a high rank $W_{ds}$) is better in \knnlm than a smaller datastore, then the hypothesis of softmax capacity is more probable.
We randomly subsample the full datastore in varying percentages and the results are shown in~\autoref{fig:subsample_dstore_only}.
The full datastore contains more than 150M entries and storing them takes 293GB when using half-precision floating points (fp16).
We can see that whether or not approximate \knn is used, the final perplexity decreases almost linearly with more percentage of the original datastore.
Even with just 5\% of the datastore size (15G), \knnlm still provides a benefit over just using the base LM.
However, even when the subsampling percentage reaches 90\%, having more entries in the datastore still provides benefits without having significant diminishing returns, suggesting that a large datastore is beneficial.

One possible reason why a larger datastore is helpful is that words can be difficult to predict.
There are several reasons: (1) They are rare, or (2) they are frequent, but they have multiple meanings and appear in different contexts.
The softmax bottleneck~\citep{yang2017breaking} suggests that the final dot product of language model $W_{sm} \cdot h_{sm}$ limits the expressivity of the output probability distributions given the context;
that is, a single output vector of a fixed (1024) size cannot express all the possible mappings between 100M training examples and 33K vocabulary outputs.
We hypothesize that \knnlm improves performance by alleviating the problem, since $W_{ds} \otimes h_{ds}$ has a higher rank and is more expressive than just $W_{sm} \cdot h_{sm}$.
In other words, \knn is a sparse approximation of the full softmax over all the embeddings in the datastore $W_{ds}$.
To test this hypothesis, we disentangle the effect of the high rank in $W_{ds}$ from the actual saved context embeddings in $W_{ds}$, by training an embedding matrix of the same desired size to test from scratch.

We explore several potential solutions for increasing the capacity of softmax, and examine if they can achieve a similar effect of \knnlm.
The first and easiest solution is to increase the embedding matrix size by adding more embedding vectors for each word type in the vocabulary.
To test this, we replace $W_{ds}$ with a much smaller matrix of size $nV \times D$, where we allocate $n$ embedding vectors for each word type.
When calculating the probability from this component, we compute the softmax over $nV$ items and sum the probabilities for each vocabulary entry to calculate the final probability.
$\text{mask-to-k}(\cdot)$ is no longer needed, as this formulation is small enough to fit the entire matrix in the GPU.
We then finetune the new $W_{ds}$ on the training data until convergence.

\begin{figure}[th]
\centering
\includegraphics[width=0.5\textwidth]{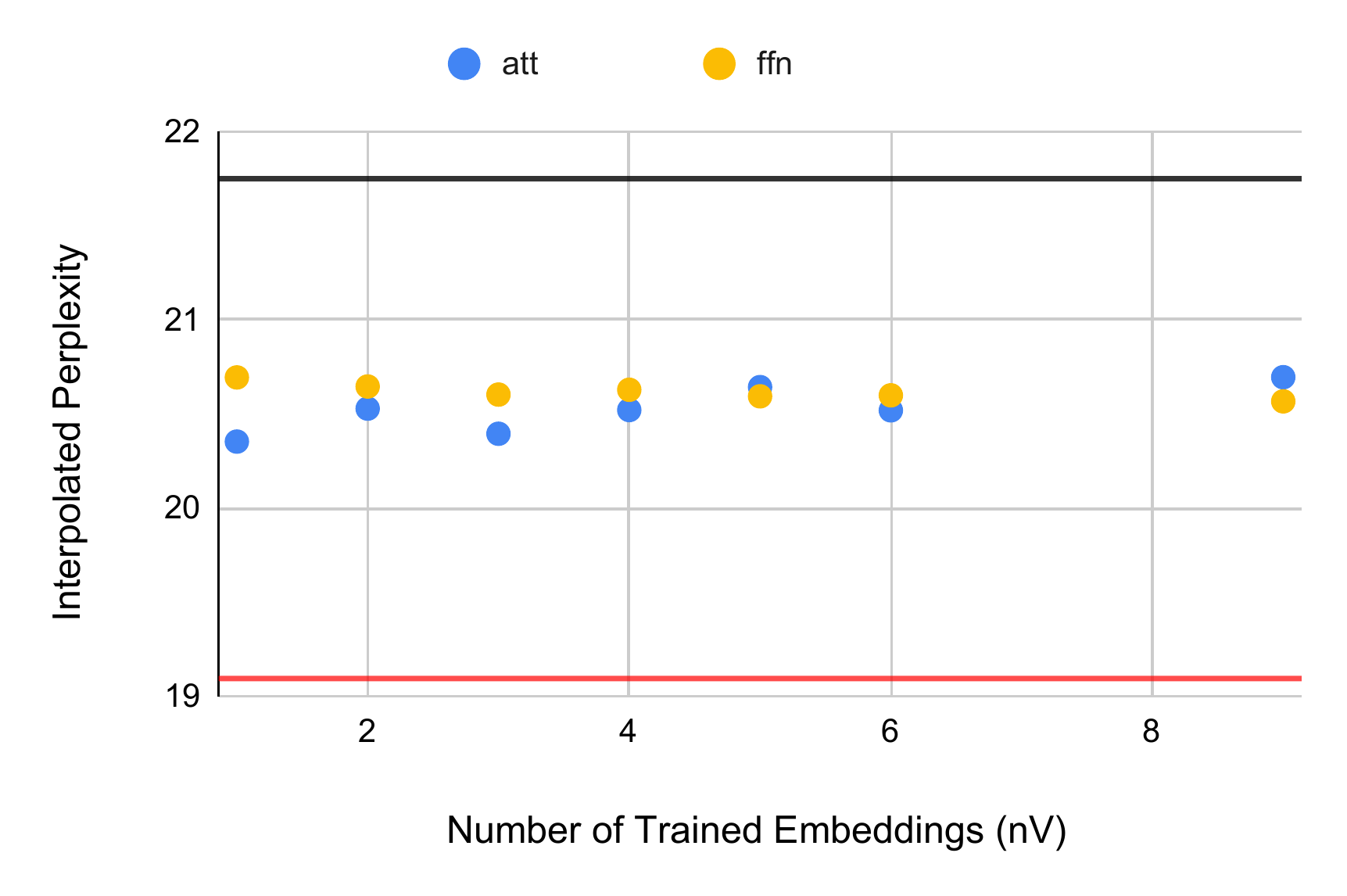}
\caption{The number of embeddings per word type ($nV$ total embeddings in $W_{ds}$) versus interpolated perplexity. The horizontal line at the top (black) represents the perplexity of the base LM. The horizontal line at the bottom (red) represents the interpolated perplexity using a full datastore with \knnlm.}
\label{fig:increasing_embedding_size}
\end{figure}

\autoref{fig:increasing_embedding_size} compares the base LM and the original \knnlm versus using either attention layer output (``att'') or feedforward layer output (``ffn'') as $h_{ds}$. 
We plot the number of embeddings for each word type ($nV$ total embeddings in $W_{ds}$) versus the interpolated perplexity, with full details found in~\autoref{app:detail_result_increasing_nv}.
In both cases, comparing with the top horizontal line which represents the perplexity of the base LM, replacing the datastore with a much smaller weight matrix (from $N_{ds}$ to $nV_{ds}$) by assigning only a few more embeddings for each word helps, although only about half as effective as \knnlm.
To give a perspective, the original datastore size is about $5000V$.
Surprisingly, we find that increasing $n$ does not always bring better performance, even though a larger datastore is better than using a small datastore in \knnlm.
We can see that when $h_{ds}=\text{ffn}$, over-parameterization provides very limited improvements, while for $h_{ds}=\text{att}$ it does not bring consistent improvements at all.
Comparing the trend of increasing the embeddings in $W_{ds}$, with the bottom horizontal line in the plot, which represents the perplexity of the standard \knnlm using the full datastore ($W_{ds}$ with approx. $5000V$ embeddings), we can see no clear trend that more trainable embeddings result in better perplexity, and that the gap between using trained embeddings and using full datastore is still significant.
This suggests that simply over-parameterizing $W_{ds}$ is not an effective method of achieving accuracy gains similar to \knnlm.

We hypothesize that this is because by just adding more embeddings, while still using the same training procedure as the original LM, the multiple embeddings for each word type after learning could still be very close to each other, and thus do not increase the softmax capacity much.
This suggests that some regularization terms may be needed during training to make the multiple embeddings not converge to the same vector, rendering over-parameterization useless.

Besides simply increasing the number of embedding vectors equally for each word type, we also propose other alternatives to increase softmax capacity.
First, we hypothesize that different word types have different difficulties for the language model to predict.
For those words that appear very frequently, they may appear in many different contexts.
As a result, instead of adding an equal number of additional embeddings to each word type, we propose to adaptively increase the number of embeddings for word types based on word frequency, or total training loss for the word.
Second, we try to break the softmax bottleneck with a Mixture of Softmax. \cite{yang2017breaking} proposes a solution to the problem using a Mixture of Softmax (MoS) to produce more linearly independent probability distributions of words given different contexts.
Last, opposite to training the word embeddings of increased size, we also consider ways to compress the datastore down to a similar-sized embedding matrix for softmax computation by clustering the whole datastore and allowing for further finetuning of the embedding matrix consisting of cluster centroids.
However, none of these alternative methods provided additional benefits over the simple multi-embedding approach.
More details on these attempts can be found in~\autoref{app:bottleneck-detail}.

\section{Approximate \knn Search \& Softmax Temperature}
\label{sec:approx_temp}

\subsection{Comparing Approximate \knn Search}
\label{sec:approximate}

To calculate $P_{kNN}$ of the non-parametric component in \autoref{eqn:general}, it is usually prohibitive to use exhaustive \knn search, and thus \citet{khandelwal2020nearest} use approximate \knn search 
using the FAISS library \citep{johnson2019billion}.
The use of FAISS (similarly to other approximate search libraries) results in two varieties of approximation.
\begin{itemize}
\item \textbf{Approximate Neighbors:} Because the search for nearest neighbors is not exact, the set of nearest neighbors might not be equivalent to the actual nearest neighbors.
Recall the function mask-to-k$(\cdot)$ in \autoref{eqn:general}, it is the function where we select the \knn entries from the datastore $W_{ds}$.
We denote ``real mask'' as the accurate nearest neighbors for mask-to-k$(\cdot)$ selection, and ``FAISS mask'' as the approximate nearest neighbors returned by the FAISS library.%
\footnote{To calculate the real mask over a large datastore, we shard the datastore into several smaller datastores, calculate the nearest neighbors for each of the smaller datastores, and combine them back together to get the final result.}
\item \textbf{Approximate Scores:} In addition, FAISS makes some approximations in calculating the distances between the query and the retrieved neighbors for efficiency purposes.
We denote ``real score'' as the scores calculated from ground truth distances between the embeddings, and ``FAISS score'' as the distances returned by FAISS approximate search.
\end{itemize}

The comparison of the different approximation settings is shown in ~\autoref{tab:knn_recompute}.
Quite surprisingly, we actually find that the interpolated perplexity with approximate search is \emph{better} than that with exact search, both with respect to the mask and the score calculation.
Intrigued by this counter-intuitive result, we explore the effect of \knn search approximation.

\begin{table}[ht]
\small
\centering
\begin{tabular}{ccHcccccc}
\toprule
 & $h_{ds}$ & $N_{ds}$ & $\otimes$ & {}+\#params      & PPL      & $\lambda$ & Interp. PPL & Oracle  \\ \midrule
Base LM         & -        & -        & -         & 0                & 21.750   & -         & -           & -      \\
\knnlm w/ FAISS mask, FAISS score        & att      & Big      & L2        & $N_{ds}\times D$ & $\infty$ & 0.271     & 19.174      & 14.230  \\
\knnlm w/ FAISS mask, real score         & att      & Big      & L2        & $N_{ds}\times D$ & $\infty$ & 0.176		     & 19.672      & 14.393       \\
\knnlm w/ real mask, real score         & att      & Big      & L2        & $N_{ds}\times D$ & $\infty$ & 0.172		     & 19.735      & 14.480       \\
\bottomrule
\end{tabular}
\caption{Performance of the parametric language model and comparison of \knnlms using the approximate versus ground truth \knn.}
\label{tab:knn_recompute}
\end{table}

\begin{figure}[th]
\centering
\includegraphics[width=0.7\textwidth]{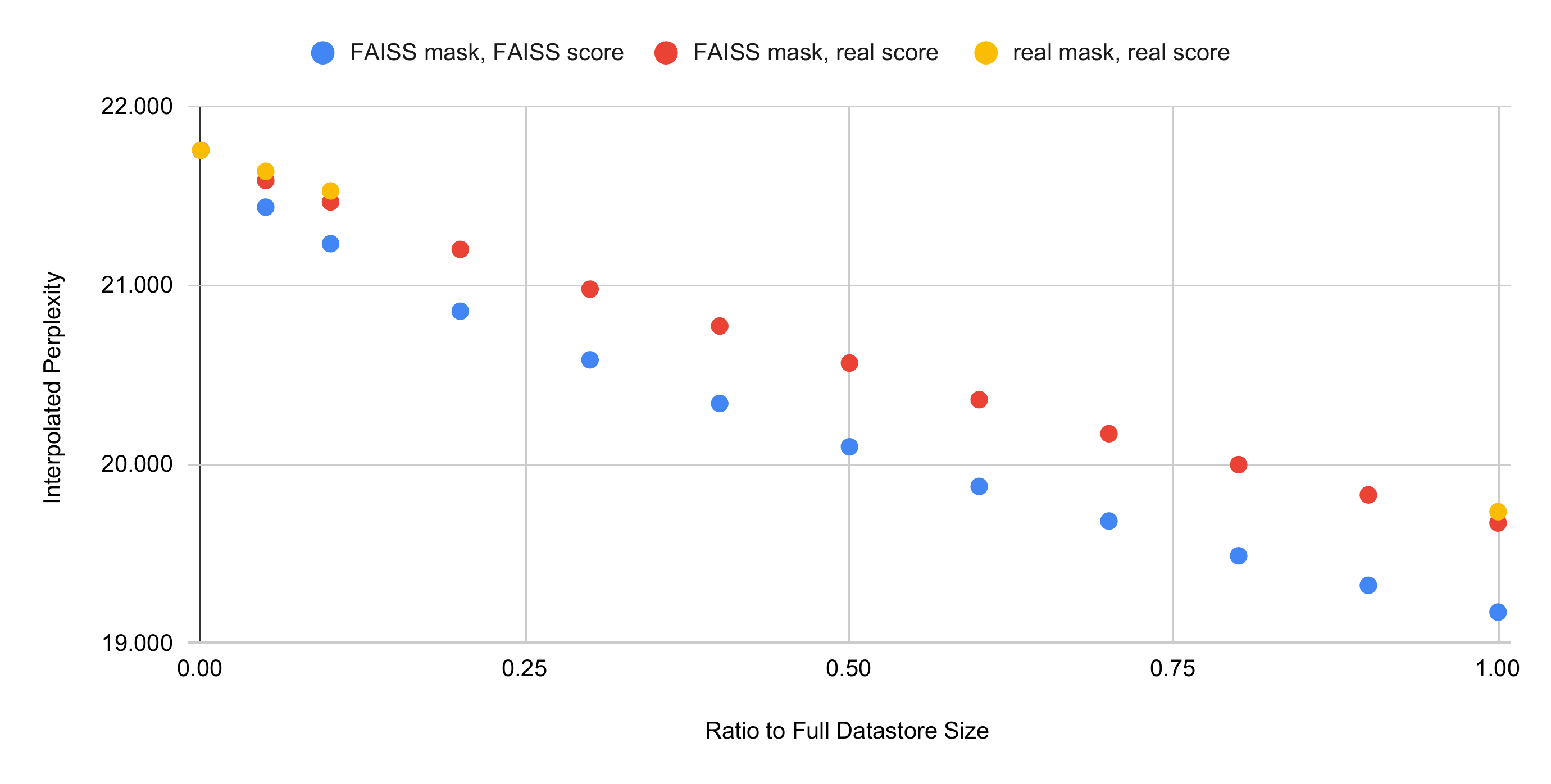}
\caption{The differences between using approximate and accurate \knn search on varying size of the datastore.}
\label{fig:datastore_size}
\end{figure}

First, we plot the subsampled size of the datastore with the interpolated perplexity~\autoref{fig:datastore_size}, a similar plot to~\autoref{fig:subsample_dstore_only}, but showcasing the comparison between approximate and real masks, between approximate and real scores in both the full datastore as well as a small subsampled datastore setting. 
We find that using an approximate FAISS mask to find nearest neighbors is better than using the ground truth nearest neighbors and that using the approximate score returned by FAISS is better than recomputing the ground truth distances between embeddings for the \knn distribution at different levels of datastore size, both at 5\% or 100\%.
Interestingly, the gap between using an approximate score or real score given the same approximate nearest neighbors (``FAISS mask, FAISS score'' vs. ``FAISS mask, real score'') is larger than that between using approximate or real nearest neighbors given the same ground truth method of calculating the distance (``real mask, real score'' vs. ``FAISS mask, real score''), for reasons we will elucidate in the next section.

\subsection{Adding Softmax Temperature to \knn Distribution}
\label{sec:temperature}

\begin{figure}[th]
 \centering
 \begin{subfigure}[b]{0.49\textwidth}
     \centering
     \includegraphics[width=\textwidth]{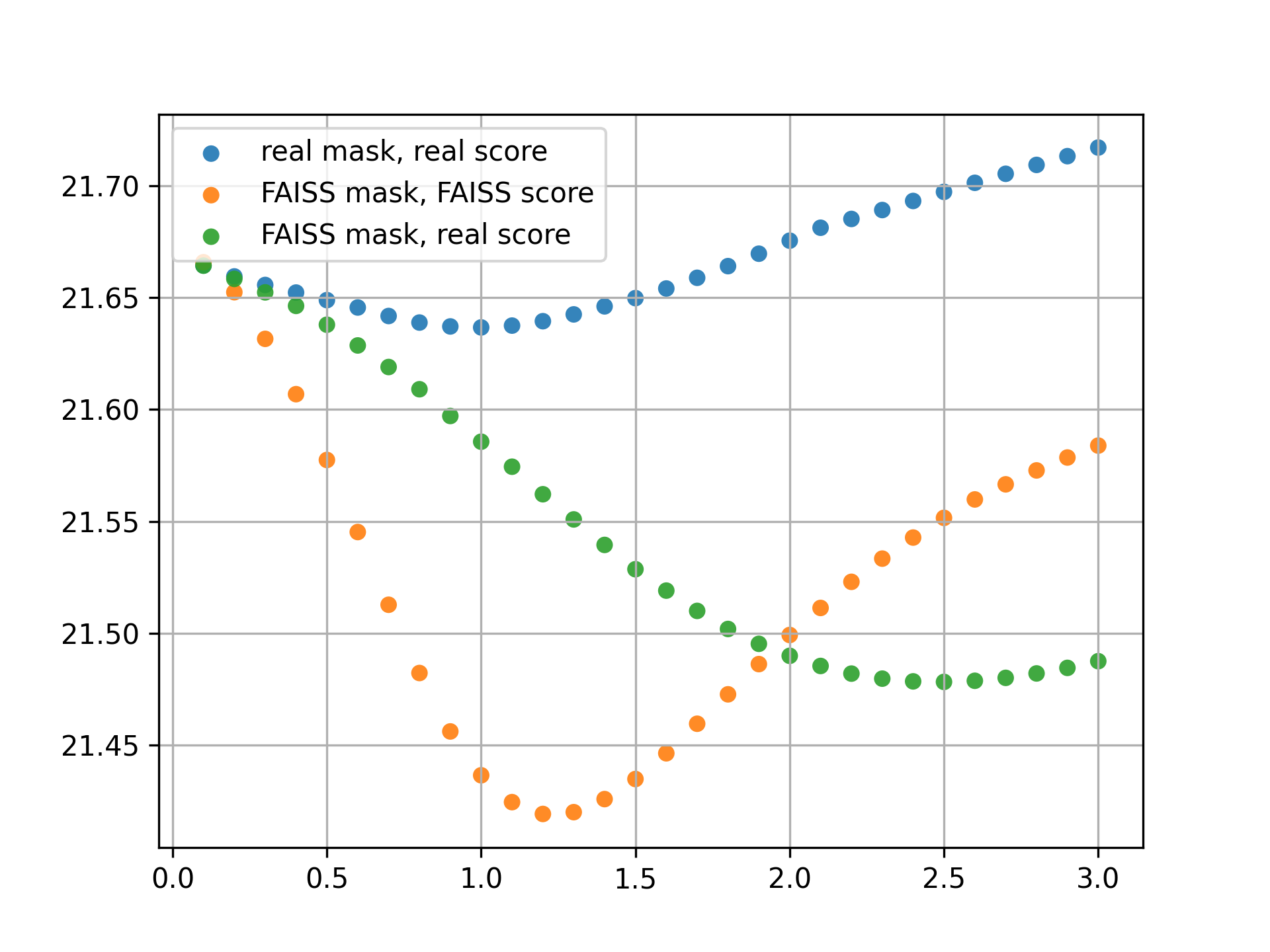}
     \caption{On 5\% subsampled datastore.}
     \label{fig:small_temperature}
 \end{subfigure}
 \hfill
 \begin{subfigure}[b]{0.49\textwidth}
     \centering
     \includegraphics[width=\textwidth]{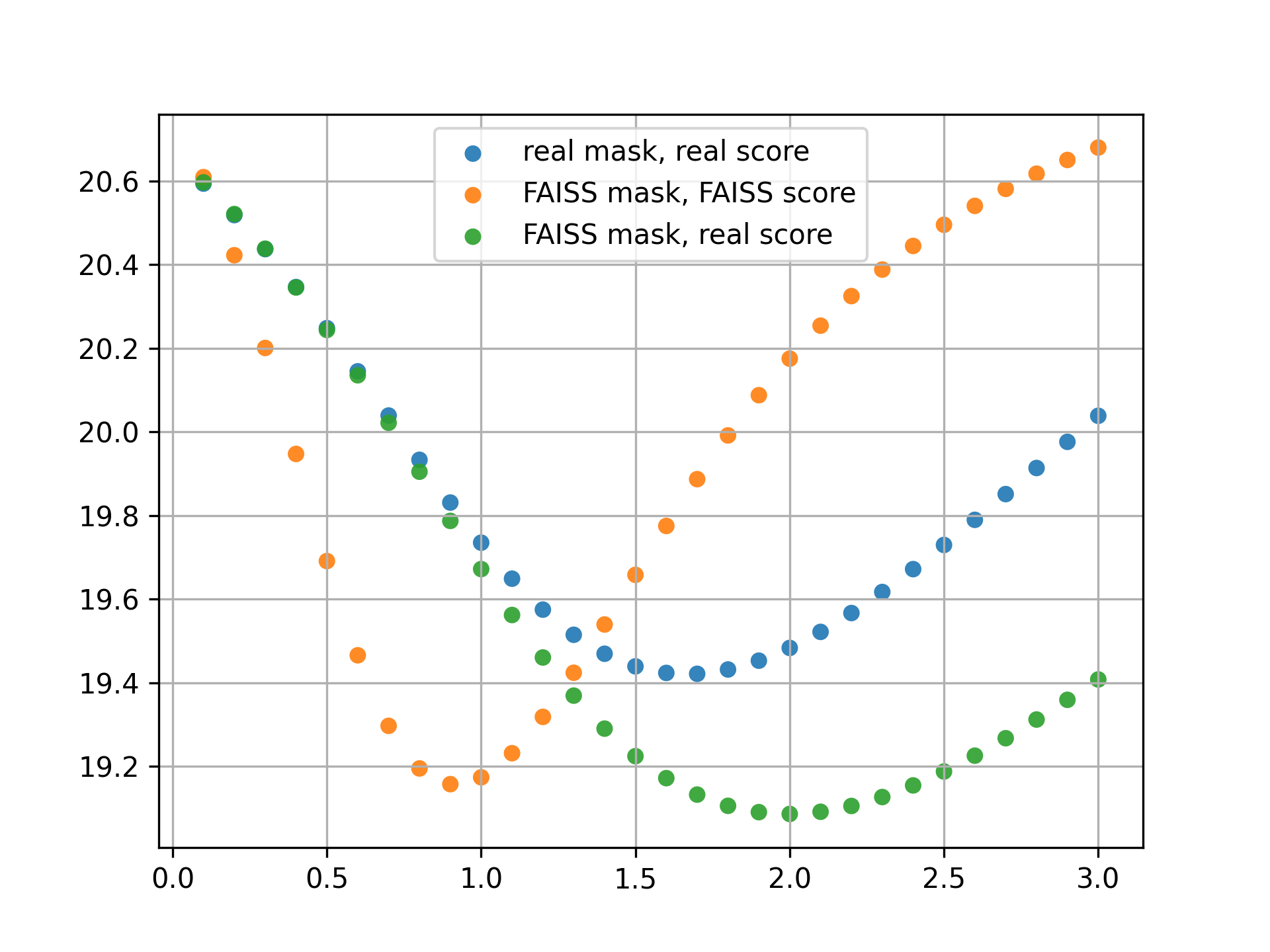}
     \caption{On full datastore.}
     \label{fig:full_temperature}
 \end{subfigure}
\caption{The interpolated perplexity varies with different softmax temperature values. }
\label{fig:temperature}
\end{figure}

Because the number of retrieved nearest neighbors, $k$ is usually much smaller than the vocabulary size $V$, intuitively, the \knn distribution $P_{kNN}$ used for interpolation tends to be more peaky than the standard LM output distribution.
When $k=1024$ and $V=33000$, as in our experiments, $P_{kNN}$ will only have a few vocabulary items with a non-zero probability.
Furthermore, many of the retrieved neighbors share the same target token and thus make the \knn distribution even peakier.
One way to control the entropy, or peakiness of the distribution is to add temperature to the logits that go into the softmax function~\citep{holtzman2019curious}.
We calculate the probability of non-parametric component $P_{kNN}$ with the following equation where $t$ is the softmax temperature:
\begin{equation}
\label{eqn:temperature}
P_{kNN} = M \text{softmax}( \text{mask-to-k}(W_{ds} \otimes h_{ds} ) / t  )
\end{equation}
In general, the higher the temperature, the less ``peaky'' the distribution would become.
We experiment with both the 5\% as well as the full datastore using different temperatures ranging from 0 to 3 at 0.1 intervals.
The results are shown in~\autoref{fig:small_temperature} and~\autoref{fig:full_temperature} respectively.

We can see that the default temperature $t=1$ does not always result in the best-interpolated perplexity and tuning softmax temperature is desirable for all sizes of datastore.
The lesson learned here is that tuning the softmax temperature for the \knn distribution is crucial for getting optimal results from each setting. 
Only coincidentally, a temperature of 1.0 was close to optimal in the original settings of \citet{khandelwal20generalization}, which hid the importance of this hyperparameter.

In both the 5\% subsampled datastore and the full datastore scenarios, temperature $t=1$ is close to optimal when using ``FAISS mask, FAISS score''.
When using either ``real mask'' or ``real score'', this is not true anymore.
Even at the optimal temperature for each setting, ``real mask, real score'' somewhat underperforms ``FAISS mask, real score''.
It is consistent with the counter-intuitive phenomenon discussed in~\autoref{sec:approximate}.

There are also differences between the two scenarios of different datastore sizes.
With the full datastore, using ``real score'' outperforms ``FAISS score'' given the same ``FAISS mask''.
However, the opposite is true when using the 5\% datastore.
This suggests that as the datastore size grows, using accurate distance values are better than the approximate ones.
The relatively small gap between using ``real score'' and ``FAISS score'' in both datastore settings shows that the main contributor to the improvements is using approximate nearest neighbors (``FAISS mask'') rather than using approximate distance values (``FAISS score'').

We hypothesize that this is related to regularization for preventing overfitting, and approximate search provides fuzziness that functions as a regularizer.
We can think of the non-parametric part in \knnlm, the \knn component as a model, where the datastore size is its model capacity, and the datastore is its training data.
Considering that the \knn component uses the exact same training data as the base parametric LM, having ground truth, accurate \knn search may cause the \knn component to overfit the training data.
Comparing the small datastore with only 5\% with the original datastore, we see that a small datastore means a small training set for the \knn ``model'' and it thus it benefits more from this regularization, both both through using the FAISS mask and FAISS score (at optimal temperature settings).
From these experiments, we can see that, surprisingly, one of the important ingredients in \knnlm seems to be approximate \knn search, which likely prevents overfitting to the datastore created from the same training set.
We further analyze this unexpected result in \autoref{app:wordanalysis}, where we find that longer words and words that appear in many different contexts have slightly better results with approximate nearest neighbors.

Coincidentally, \citet{he2021efficient} find that dimensionality reduction using PCA on datastore vectors (from 1024 to 512 dimensions) improves the perplexity of the original \knnlm from 16.46 to 16.25, which can be explained by our findings as PCA may provide another source of ``approximation'' that contributes to regularization.

Notably, similar effects, where an approximation component lead to better generalization, have been reported in other NLP tasks as well, and are sometimes referred to as ``beneficial search bias'', when modeling errors cause the highest-scoring solution to not be the correct one:
\citet{meister2020best} suggest that ``quite surprisingly, beam search often returns better results than exact inference due to beneficial search bias for NLP tasks.'' 
\citet{stahlberg2019nmt} also conclude that ``vanilla NMT in its current form requires just the right amount of beam search errors, which, from a modeling perspective, is a highly unsatisfactory conclusion indeed, as the model often prefers an empty translation''.

\section{Probably Wrong Hypotheses for Why \knnlms Work}

The results in the previous sections are the result of extensive analysis and experimentation, in which we also tested a number of hypotheses that did \emph{not} turn out to have a significant effect.
Additional details of these hypotheses are detailed in~\autoref{app:failed}, and we hope that they may provide ideas for future improvements of retrieval-based LMs.

\paragraph{Ensemble of Distance Metrics} 
We hypothesized that the ensemble of two distance metrics: the standard inner product distance (which the LM uses) and the L2 distance (which the \knn component uses), is the key to the improvement. 
However, we found that similar gains can be achieved using the inner-product metric for the retrieved \knn. More details can be found in~\autoref{app:dist_metric}.

\paragraph{Ensembling of Two Models}
We hypothesized that the \knn component merely provides another model for ensembling. The improvement from \knnlm is \emph{purely} due to the ensembling effect of different models.
However, we found that \knnlm's improvement is orthogonal to ensembling with a different base LM.
More details can be found in~\autoref{app:ensembling}.

\paragraph{Sparsification} 
The $\text{mask-to-k}(\cdot)$ used by \knn retrieval induces sparsity in the distribution over the vocabulary, due to a small $k$ (typically 1024) compared to the size of the vocabulary $V$ (33K in our experiments and 260K in the original settings of \citet{khandelwal20generalization}).
We hypothesized that \knnlm increases the probability of the top-$k$ entries while taking ``probability mass'' from the long tail of unlikely word types. 
However, we could not gain any benefits solely from sparsifying the output probability of a standard LM and interpolating it with the original LM.
More details can be found in~\autoref{app:sparsification}.


\paragraph{Stolen Probabilities}
The \emph{stolen probabilities} effect \citep{demeter2020stolen} refers to the situation where the output embeddings of an LM are learned such that some words are geometrically placed \emph{inside} the convex hull that is formed by other word embeddings and can thus never be ``selected'' as the argmax word.
We hypothesized that \knnlm solves the stolen probabilities problem by allowing to assign the highest probability to \emph{any} word, given a test context that is close enough to that word's datastore key. 
However, we found that \emph{none} of the vectors in our embedding matrix and in the original embedding matrix of \citet{khandelwal20generalization} is located in the convex hull of the others, which is consistent with the findings of \citet{grivas2022low}.
More details can be found in~\autoref{app:stolen_prob}.


\paragraph{Memorization}
We hypothesized that the \knn component simply provides memorization of the training set. However, we could not improve a standard LM by interpolating its probability with another standard LM that was further trained to overfit the training set.
More details can be found in~\autoref{app:overfitting}.

\paragraph{Soft Labels}
We hypothesized that \knnlm's improvement lies in reducing the ``over-correction'' error when training with 1-hot labels, as hypothesized by \citet{yang2022nearest}, and that retrieving neighbors is not important.
If only ``soft labels'' are the key, we could hypothetically improve the performance of another fresh LM with the same model architecture but trained with the soft labels from the base LM, instead of from \knnlm.
This separates the effect of ``soft labeling'' from the additional guidance provided by \knn.
However, this does not help with the interpolated perplexity at all.
More details can be found in~\autoref{app:soft_label}.

\paragraph{Optimizing Interpolated Loss}
We hypothesized that the standard LM cross-entropy training loss does not emphasize the examples where base LM performs badly which could benefit from \knn, and directly optimizing the interpolated loss of standard LM and a separate trainable softmax layer could be a better alternative.
However, we could not gain any benefits by training an additional softmax layer together with a base LM using the interpolated loss.
More details can be found in~\autoref{app:interpolated_loss}.

\section{Conclusion}
In this paper, we investigate why \knnlm improves perplexity, even when retrieving examples from the same training data that the base LM was trained on.
By proposing and testing various hypotheses and performing extensive ablation studies, we find that the key to \knnlm's success is threefold:
\begin{enumerate}
\item Ensembling different input representations -- the feedforward layer output and the attention layer output -- can recover 55\% of the performance, even without retrieval.
\item One of the most unexpected discoveries in the paper is that
using \emph{approximate} nearest neighbor search allows \knnlms to generalize better than exact nearest neighbor search, possibly due to a regularization effect.
\item Tuning the softmax temperature for the \knn distribution is crucial to adjust the standard LM output distribution with the distribution created by the retrieved neighbors{}' distances.
\end{enumerate}

We performed extensive experiments which ruled out other hypotheses as to why \knnlms work,
such as over-parameterization, datastore clustering, sparsification, overfitting, ensembling of distance metrics, and alternative training methods.

We believe that this work unlocks a variety of exciting research directions for efficient \knn alternatives.
For example, exploring methods that replace the \knn component with trainable parameters and achieve comparable results without the latency burden of \knnlm.

\section*{Acknowledgement}
We thank Ramesh Nallapati, Sudipta Sengupta, Dan Roth, Daniel Fried and Xiaosen Zheng for the helpful discussions and feedback. 
This project was supported by a gift from AWS AI. 
Frank F. Xu is supported by IBM Ph.D. Fellowship.

\bibliography{main}

\begin{thebibliography}{35}
\providecommand{\natexlab}[1]{#1}
\providecommand{\url}[1]{\texttt{#1}}
\expandafter\ifx\csname urlstyle\endcsname\relax
  \providecommand{\doi}[1]{doi: #1}\else
  \providecommand{\doi}{doi: \begingroup \urlstyle{rm}\Url}\fi

\bibitem[Alon et~al.(2022)Alon, Xu, He, Sengupta, Roth, and
  Neubig]{alon2022neuro}
Uri Alon, Frank~F Xu, Junxian He, Sudipta Sengupta, Dan Roth, and Graham
  Neubig.
\newblock Neuro-symbolic language modeling with automaton-augmented retrieval.
\newblock \emph{arXiv preprint arXiv:2201.12431}, 2022.

\bibitem[Ba et~al.(2016)Ba, Kiros, and Hinton]{ba2016layer}
Jimmy~Lei Ba, Jamie~Ryan Kiros, and Geoffrey~E Hinton.
\newblock Layer normalization.
\newblock \emph{arXiv preprint arXiv:1607.06450}, 2016.

\bibitem[Baevski and Auli(2018)]{baevski2018adaptive}
Alexei Baevski and Michael Auli.
\newblock Adaptive input representations for neural language modeling.
\newblock \emph{arXiv preprint arXiv:1809.10853}, 2018.

\bibitem[Bengio et~al.(2003)Bengio, Ducharme, Vincent, and
  Jauvin]{bengio2003neural}
Yoshua Bengio, R{\'e}jean Ducharme, Pascal Vincent, and Christian Jauvin.
\newblock A neural probabilistic language model.
\newblock \emph{Journal of machine learning research}, 3\penalty0
  (Feb):\penalty0 1137--1155, 2003.

\bibitem[Borgeaud et~al.(2022)Borgeaud, Mensch, Hoffmann, Cai, Rutherford,
  Millican, Van Den~Driessche, Lespiau, Damoc, Clark,
  et~al.]{borgeaud2022improving}
Sebastian Borgeaud, Arthur Mensch, Jordan Hoffmann, Trevor Cai, Eliza
  Rutherford, Katie Millican, George~Bm Van Den~Driessche, Jean-Baptiste
  Lespiau, Bogdan Damoc, Aidan Clark, et~al.
\newblock Improving language models by retrieving from trillions of tokens.
\newblock In \emph{International conference on machine learning}, pages
  2206--2240. PMLR, 2022.

\bibitem[Brown et~al.(2020)Brown, Mann, Ryder, Subbiah, Kaplan, Dhariwal,
  Neelakantan, Shyam, Sastry, Askell, et~al.]{brown2020language}
Tom~B Brown, Benjamin Mann, Nick Ryder, Melanie Subbiah, Jared Kaplan, Prafulla
  Dhariwal, Arvind Neelakantan, Pranav Shyam, Girish Sastry, Amanda Askell,
  et~al.
\newblock Language models are few-shot learners.
\newblock \emph{arXiv preprint arXiv:2005.14165}, 2020.

\bibitem[Clauset et~al.(2009)Clauset, Shalizi, and Newman]{clauset2009power}
Aaron Clauset, Cosma~Rohilla Shalizi, and Mark~EJ Newman.
\newblock Power-law distributions in empirical data.
\newblock \emph{SIAM review}, 51\penalty0 (4):\penalty0 661--703, 2009.

\bibitem[Demeter et~al.(2020)Demeter, Kimmel, and Downey]{demeter2020stolen}
David Demeter, Gregory Kimmel, and Doug Downey.
\newblock Stolen probability: A structural weakness of neural language models.
\newblock In \emph{Proceedings of the 58th Annual Meeting of the Association
  for Computational Linguistics}, pages 2191--2197, 2020.

\bibitem[Devlin et~al.(2018)Devlin, Chang, Lee, and Toutanova]{devlin2018bert}
Jacob Devlin, Ming-Wei Chang, Kenton Lee, and Kristina Toutanova.
\newblock B{ERT}: Pre-training of deep bidirectional transformers for language
  understanding.
\newblock \emph{arXiv preprint arXiv:1810.04805}, 2018.

\bibitem[Grave et~al.(2017)Grave, Ciss{\'e}, and Joulin]{grave2017unbounded}
Edouard Grave, Moustapha Ciss{\'e}, and Armand Joulin.
\newblock Unbounded cache model for online language modeling with open
  vocabulary.
\newblock \emph{arXiv preprint arXiv:1711.02604}, 2017.

\bibitem[Grivas et~al.(2022)Grivas, Bogoychev, and Lopez]{grivas2022low}
Andreas Grivas, Nikolay Bogoychev, and Adam Lopez.
\newblock Low-rank softmax can have unargmaxable classes in theory but rarely
  in practice.
\newblock In \emph{Proceedings of the 60th Annual Meeting of the Association
  for Computational Linguistics (Volume 1: Long Papers)}, pages 6738--6758,
  2022.

\bibitem[Guu et~al.(2018)Guu, Hashimoto, Oren, and Liang]{guu2018generating}
Kelvin Guu, Tatsunori~B Hashimoto, Yonatan Oren, and Percy Liang.
\newblock Generating sentences by editing prototypes.
\newblock \emph{Transactions of the Association for Computational Linguistics},
  6:\penalty0 437--450, 2018.

\bibitem[He et~al.(2020)He, Berg-Kirkpatrick, and Neubig]{he2020learning}
Junxian He, Taylor Berg-Kirkpatrick, and Graham Neubig.
\newblock Learning sparse prototypes for text generation.
\newblock \emph{arXiv preprint arXiv:2006.16336}, 2020.

\bibitem[He et~al.(2021)He, Neubig, and Berg-Kirkpatrick]{he2021efficient}
Junxian He, Graham Neubig, and Taylor Berg-Kirkpatrick.
\newblock Efficient nearest neighbor language models.
\newblock \emph{arXiv preprint arXiv:2109.04212}, 2021.

\bibitem[Hinton et~al.(2015)Hinton, Vinyals, Dean,
  et~al.]{hinton2015distilling}
Geoffrey Hinton, Oriol Vinyals, Jeff Dean, et~al.
\newblock Distilling the knowledge in a neural network.
\newblock \emph{arXiv preprint arXiv:1503.02531}, 2\penalty0 (7), 2015.

\bibitem[Holtzman et~al.(2019)Holtzman, Buys, Du, Forbes, and
  Choi]{holtzman2019curious}
Ari Holtzman, Jan Buys, Li~Du, Maxwell Forbes, and Yejin Choi.
\newblock The curious case of neural text degeneration.
\newblock \emph{arXiv preprint arXiv:1904.09751}, 2019.

\bibitem[Johnson et~al.(2019)Johnson, Douze, and J{\'e}gou]{johnson2019billion}
Jeff Johnson, Matthijs Douze, and Herv{\'e} J{\'e}gou.
\newblock Billion-scale similarity search with {GPUs}.
\newblock \emph{IEEE Transactions on Big Data}, 7\penalty0 (3):\penalty0
  535--547, 2019.

\bibitem[Joulin et~al.(2017)Joulin, Ciss{\'e}, Grangier, J{\'e}gou,
  et~al.]{joulin2017efficient}
Armand Joulin, Moustapha Ciss{\'e}, David Grangier, Herv{\'e} J{\'e}gou, et~al.
\newblock Efficient softmax approximation for gpus.
\newblock In \emph{International conference on machine learning}, pages
  1302--1310. PMLR, 2017.

\bibitem[Khandelwal et~al.(2020{\natexlab{a}})Khandelwal, Fan, Jurafsky,
  Zettlemoyer, and Lewis]{khandelwal2020nearest}
Urvashi Khandelwal, Angela Fan, Dan Jurafsky, Luke Zettlemoyer, and Mike Lewis.
\newblock Nearest neighbor machine translation.
\newblock \emph{arXiv preprint arXiv:2010.00710}, 2020{\natexlab{a}}.

\bibitem[Khandelwal et~al.(2020{\natexlab{b}})Khandelwal, Levy, Jurafsky,
  Zettlemoyer, and Lewis]{khandelwal20generalization}
Urvashi Khandelwal, Omer Levy, Dan Jurafsky, Luke Zettlemoyer, and Mike Lewis.
\newblock {Generalization through Memorization: Nearest Neighbor Language
  Models}.
\newblock In \emph{International Conference on Learning Representations
  (ICLR)}, 2020{\natexlab{b}}.

\bibitem[Liu et~al.(2019)Liu, Ott, Goyal, Du, Joshi, Chen, Levy, Lewis,
  Zettlemoyer, and Stoyanov]{liu2019roberta}
Yinhan Liu, Myle Ott, Naman Goyal, Jingfei Du, Mandar Joshi, Danqi Chen, Omer
  Levy, Mike Lewis, Luke Zettlemoyer, and Veselin Stoyanov.
\newblock Roberta: A robustly optimized bert pretraining approach.
\newblock \emph{arXiv preprint arXiv:1907.11692}, 2019.

\bibitem[Meister et~al.(2020{\natexlab{a}})Meister, Salesky, and
  Cotterell]{meister2020generalized}
Clara Meister, Elizabeth Salesky, and Ryan Cotterell.
\newblock Generalized entropy regularization or: There's nothing special about
  label smoothing.
\newblock \emph{arXiv preprint arXiv:2005.00820}, 2020{\natexlab{a}}.

\bibitem[Meister et~al.(2020{\natexlab{b}})Meister, Vieira, and
  Cotterell]{meister2020best}
Clara Meister, Tim Vieira, and Ryan Cotterell.
\newblock Best-first beam search.
\newblock \emph{Transactions of the Association for Computational Linguistics},
  8:\penalty0 795--809, 2020{\natexlab{b}}.

\bibitem[Merity et~al.(2016)Merity, Xiong, Bradbury, and
  Socher]{merity2016pointer}
Stephen Merity, Caiming Xiong, James Bradbury, and Richard Socher.
\newblock Pointer sentinel mixture models.
\newblock \emph{arXiv preprint arXiv:1609.07843}, 2016.

\bibitem[Merity et~al.(2018)Merity, Keskar, and Socher]{merity2018regularizing}
Stephen Merity, Nitish~Shirish Keskar, and Richard Socher.
\newblock Regularizing and optimizing {LSTM} language models.
\newblock In \emph{Proceedings of ICLR}, 2018.

\bibitem[Ney et~al.(1994)Ney, Essen, and Kneser]{ney1994structuring}
Hermann Ney, Ute Essen, and Reinhard Kneser.
\newblock On structuring probabilistic dependences in stochastic language
  modelling.
\newblock \emph{Computer Speech \& Language}, 8\penalty0 (1):\penalty0 1--38,
  1994.

\bibitem[Pereyra et~al.(2017)Pereyra, Tucker, Chorowski, Kaiser, and
  Hinton]{pereyra2017regularizing}
Gabriel Pereyra, George Tucker, Jan Chorowski, {\L}ukasz Kaiser, and Geoffrey
  Hinton.
\newblock Regularizing neural networks by penalizing confident output
  distributions.
\newblock \emph{arXiv preprint arXiv:1701.06548}, 2017.

\bibitem[Radford et~al.(2019)Radford, Wu, Child, Luan, Amodei, Sutskever,
  et~al.]{radford2019language}
Alec Radford, Jeffrey Wu, Rewon Child, David Luan, Dario Amodei, Ilya
  Sutskever, et~al.
\newblock Language models are unsupervised multitask learners.
\newblock \emph{OpenAI blog}, 1\penalty0 (8):\penalty0 9, 2019.

\bibitem[Sanh et~al.(2019)Sanh, Debut, Chaumond, and Wolf]{sanh2019distilbert}
Victor Sanh, Lysandre Debut, Julien Chaumond, and Thomas Wolf.
\newblock Distilbert, a distilled version of bert: smaller, faster, cheaper and
  lighter.
\newblock \emph{arXiv preprint arXiv:1910.01108}, 2019.

\bibitem[Sennrich et~al.(2015)Sennrich, Haddow, and Birch]{sennrich2015neural}
Rico Sennrich, Barry Haddow, and Alexandra Birch.
\newblock Neural machine translation of rare words with subword units.
\newblock \emph{arXiv preprint arXiv:1508.07909}, 2015.

\bibitem[Stahlberg and Byrne(2019)]{stahlberg2019nmt}
Felix Stahlberg and Bill Byrne.
\newblock On nmt search errors and model errors: Cat got your tongue?
\newblock \emph{arXiv preprint arXiv:1908.10090}, 2019.

\bibitem[Szegedy et~al.(2016)Szegedy, Vanhoucke, Ioffe, Shlens, and
  Wojna]{szegedy2016rethinking}
Christian Szegedy, Vincent Vanhoucke, Sergey Ioffe, Jon Shlens, and Zbigniew
  Wojna.
\newblock Rethinking the inception architecture for computer vision.
\newblock In \emph{Proceedings of the IEEE conference on computer vision and
  pattern recognition}, pages 2818--2826, 2016.

\bibitem[Wang et~al.(2022)Wang, Fan, Chen, and Xiong]{Wang2022EfficientCK}
Dexin Wang, Kai Fan, Boxing Chen, and Deyi Xiong.
\newblock Efficient cluster-based k-nearest-neighbor machine translation.
\newblock \emph{ArXiv}, abs/2204.06175, 2022.

\bibitem[Yang et~al.(2017)Yang, Dai, Salakhutdinov, and
  Cohen]{yang2017breaking}
Zhilin Yang, Zihang Dai, Ruslan Salakhutdinov, and William~W Cohen.
\newblock Breaking the softmax bottleneck: A high-rank rnn language model.
\newblock \emph{arXiv preprint arXiv:1711.03953}, 2017.

\bibitem[Yang et~al.(2022)Yang, Sun, and Wan]{yang2022nearest}
Zhixian Yang, Renliang Sun, and Xiaojun Wan.
\newblock Nearest neighbor knowledge distillation for neural machine
  translation.
\newblock In \emph{Proceedings of the 2022 Conference of the North American
  Chapter of the Association for Computational Linguistics: Human Language
  Technologies}, pages 5546--5556, Seattle, United States, July 2022.
  Association for Computational Linguistics.
\newblock \doi{10.18653/v1/2022.naacl-main.406}.
\newblock URL \url{https://aclanthology.org/2022.naacl-main.406}.

\end{thebibliography}
\newpage
\appendix
\section{\knnlm Generalization to Other LMs}
\label{app:generalization-knnlm}

\begin{table}[h]
\small
\centering
\begin{tabular}{ccccc}
\toprule
 & \#params     &  Base LM PPL  &  \knnlm PPL  & Absolute PPL Gain \\ \midrule
Ours         & 268M & 21.75 & 19.17 & 2.58 \\ \midrule
Distilled-GPT2         & 82M & 18.25 & 14.84 & 3.41 \\
GPT2-small         & 117M       & 14.84      & 12.55 & 2.29\\
GPT2-medium         & 345M  & 11.55 & 10.37 & 1.18 \\
GPT2-large         & 774M  & 10.56 & 9.76 & 0.80 \\
\bottomrule
\end{tabular}
\caption{Performance of \knnlm applied to other pretrained language models of different sizes.}
\label{tab:knn_other_models}
\end{table}

To test the generalizability of \knnlm, we follow the same experimental setup as used in \autoref{sec:baseline}.
We select several pretrained models from the GPT2 family~\citep{radford2019language} of various parameter counts, plus a distilled version of GPT2, DistillGPT2.~\citep{sanh2019distilbert}
We take the pretrained model checkpoint, build the datastore and evaluate on the Wikitext-103 dataset splits. 
The results are shown in~\autoref{tab:knn_other_models}.
We can see that \knnlms has good generalizability on other models.
It improves the perplexity of all the base LMs tested.
However, the larger the model is, and usually the better the base LM's perplexity is, the less gain can be achieved from adding \knn.
Note that our model is trained from scratch on Wikitext-103 dataset and thus even with a relatively large model size, the perplexity and perplexity gain from adding \knn is still less than models with pretraining.
Without loss of generalizability, we will use our own trained-from-scratch model as the base LM in the following sections for ablation study.

\section{Detailed Results for Increasing the Softmax Capacity}
\label{app:detail_result_increasing_nv}

\begin{table}[th]
\centering
\small
\begin{tabular}{HcccccHccHH}
\toprule
Prev. Layers & $h_{ds}$ & $N_{ds}$ & $\otimes$ & {}+\#params      & PPL      & $\lambda$ & Interp. & Oracle & Overlap & Helping \\ \midrule
same         & -        & -        & -         & 0                & 21.750   & -         & -           & -      & -       & -       \\\midrule
same         & att      & Big      & IP        & $N_{ds}\times D$ & $\infty$ & 0.266     & 19.095      & 14.077 & 0.890   & 0.402   \\
same         & att      & 1x       & IP        & $V\times D$      & 22.584   & 0.407     & 20.353      & 16.954 & 0.541   & 0.483   \\
same         & att      & 2x       & IP        & $2V\times D$     & 21.903   & 0.437     & 20.529      & 17.432 & 0.520   & 0.502   \\
same         & att      & 3x       & IP        & $3V\times D$     & 22.434   & 0.417     & 20.395      & 17.132 & 0.529   & 0.496   \\
same         & att      & 4x       & IP        & $4V\times D$     & 21.936   & 0.437     & 20.521      & 17.423 & 0.520   & 0.504   \\
same         & att      & 5x       & IP        & $5V\times D$     & 22.025   & 0.427     & 20.643      & 17.560 & 0.517   & 0.502   \\
same         & att      & 6x       & IP        & $6V\times D$     & 21.972   & 0.437     & 20.519      & 17.422 & 0.519   & 0.505   \\ 
same         & att      & 9x       & IP        & $9V\times D$     & 22.084   & 0.417     & 20.696      & 17.631 & 0.517   & 0.504   \\ \midrule
same         & ffn      & Big      & IP        & $N_{ds}\times D$ & $\infty$ & 0.050     & 21.101      & 16.254 & 0.586   & 0.340   \\
same         & ffn      & 1x       & IP        & $V\times D$      & 20.920   & 0.607     & 20.694      & 18.772 & 0.536   & 0.432   \\
same         & ffn      & 2x       & IP        & $2V\times D$     & 20.889   & 0.607     & 20.646      & 18.701 & 0.538   & 0.430   \\
same         & ffn      & 3x       & IP        & $3V\times D$     & 20.829   & 0.622     & 20.603      & 18.717 & 0.538   & 0.439   \\
same         & ffn      & 4x       & IP        & $4V\times D$     & 20.769   & 0.688     & 20.629      & 18.876 & 0.542   & 0.452   \\
same         & ffn      & 5x       & IP        & $5V\times D$     & 20.720   & 0.703     & 20.594      & 18.878 & 0.545   & 0.459   \\
same         & ffn      & 6x       & IP        & $6V\times D$     & 20.726   & 0.708     & 20.599      & 18.902 & 0.548   & 0.469   \\
same         & ffn      & 9x       & IP        & $9V\times D$     & 20.687   & 0.718     & 20.567      & 18.887 & 0.548   & 0.476   \\ \bottomrule
\end{tabular}
\caption{Performance comparison of \knn baselines and models with learnable embeddings as datastore alternative. $h_{ds}$ is either attention layer output (att) or feedforward layer output (ffn).}
\label{tab:softmax_bottleneck}
\end{table}

\section{Alternative Methods for Increasing Softmax Capacity}
\label{app:bottleneck-detail}

\subsection{Adaptive Increasing Embedding Size}
We hypothesize that different word types have different difficulties for the language model to predict.
For those words that appear very frequently, they may appear in many different contexts.
As a result, instead of adding equal number of additional embeddings to each word type, we propose to adaptively increase the number of embeddings for word types based on word frequency, or total training loss for the word.
Based on the intuition of Zipf's law~\citep{clauset2009power}, we assign $1+\log_{b}{f_v}$ for each word type $v\in V$, based on either the frequency or the total training loss of the word, $f_v$.
The $b$ is a hyperparameter that could be tuned.
To ensure fair comparison, we tune $b$ so that for each experiment the total number of embeddings matches: $\sum_{v\in V} 1+\log_{b}{f_v} = nV$.
The results are shown in~\autoref{tab:adaptive_results}.
We can see that although nice in paper, given the same number of total embeddings, adaptively increasing the number of embeddings assigned for each word type does not make a significant difference in the final perplexity, when compared with the models that use equal number of embeddings for each word type.

\begin{table}[h]
\centering
\small
\begin{tabular}{cccccccccHH}
\toprule
                  & $h_{ds}$ & $N_{ds}$ & $\otimes$ & {}+\#params                  & PPL      & $\lambda$ & Interp. PPL & Oracle & Overlap & Helping \\ \midrule
Base LM        & -        & -   & -         & 0                & 21.750   & -         & -           & -      & -       & -       \\
KNN            & att      & Big   & L2        & $N_{ds}\times D$ & $\infty$ & 0.271     & 19.174      & 14.230 & 1.000   & 0.380   \\
KNN            & att      & Big   & IP        & $N_{ds}\times D$ & $\infty$ & 0.266     & 19.095      & 14.077 & 0.890   & 0.402   \\
Equal Per Word & att      & 3x  & IP        & $3V\times D$           & 22.434   & 0.417     & 20.395      & 17.132 & 0.529   & 0.496   \\
Loss Weighted  & att      & 3x  & IP        & $3V\times D$           & 21.948   & 0.437     & 20.440      & 17.303 & 0.522   & 0.503   \\
Freq. Weighted & att      & 3x  & IP        & $3V\times D$           & 22.507   & 0.412     & 20.387      & 17.105 & 0.532   & 0.495   \\
KNN            & ffn      &Big   & L2        & $N_{ds}\times D$ & $\infty$ & 0.065     & 20.734      & 15.594 & 0.600   & 0.357   \\
KNN            & ffn      & Big   & IP        & $N_{ds}\times D$ & $\infty$ & 0.050     & 21.101      & 16.254 & 0.586   & 0.340   \\
Equal Per Word & ffn      & 3x  & IP        & $3V\times D$           & 20.829   & 0.622     & 20.603      & 18.717 & 0.538   & 0.439   \\
Loss Weighted  & ffn      & 3x  & IP        & $3V\times D$          & 20.764   & 0.713     & 20.659      & 18.978 & 0.543   & 0.465   \\
Freq. Weighted & ffn      & 3x  & IP        & $3V\times D$          & 20.757   & 0.658     & 20.572      & 18.782 & 0.547   & 0.452 \\ 
 \bottomrule
\end{tabular}
\caption{Performance comparison of \knn baselines and several configurations that adaptively increase the embedding size with training loss or word frequency.}
\label{tab:adaptive_results}
\end{table}

\subsection{Mixture of Softmaxes}
\cite{yang2017breaking} proposes a solution to the problem using a Mixture of Softmax (MoS) to produce more linearly independent probability distributions of words given different contexts.
Suppose that there are a total of $R$ mixture components. 
MoS first uses $R$ linear layers with weight $w_r$ to transform the current query context vector $h_{ds}$ into $w_rh_{ds}$.
With a shared word embedding matrix $W_{sm}$, we can calculate each softmax component's probability distribution with $\text{softmax}(W_{sm} \cdot w_rh_{ds})$.
The mixture distribution is then given by:
\begin{equation}
    P_{MoS} = \sum_r^R \pi_{r,h_{ds}}\text{softmax}(W_{sm} \cdot w_rh_{ds})
\end{equation}
The prior weights are calculated using another linear layer with weight $w_{\pi}$, as $\pi_{r,h_{ds}} = \text{softmax}(w_{\pi}h_{ds})$.
The softmax ensures that $\sum_r^R \pi_{r,h_{ds}} = 1$.
Comparing the MoS with the first term in \autoref{eqn:general}, $M \text{softmax}( \text{mask-to-k}(W_{ds} \otimes h_{ds} ) )$, we can see that there are some connections between the two.
MoS eliminates the $\text{mask-to-k}(\cdot)$ operation, and replaces the single softmax across a very large vector (size of datastore), into multiple smaller softmaxes, each across only a vector of the size of vocabulary.
As a result, the huge $W_{ds}$ is replaced by several linear layers to project the word embedding matrix.
Now the first term becomes:
\begin{align}
    &M (\oplus_r^R\text{softmax}(W_{sm} \cdot w_rh_{ds}))\\
    &M_{ir} = \pi_{r,h_{ds}}, \forall i \leq V
\end{align}
where $\oplus$ represents the vector concatenation operation, and the aggregation matrix $M$ now contains the mixture weights for each softmax being concatenated.
We perform experiments with a varying number of mixtures ($R$), different definitions $h_{ds}$, and whether to finetune the output word embeddings $W_{sm}$.
We allow finetuning the word embedding when we use attention layer output as context vector, since the word embedding matrix is trained with feedforward layer output originally.
The results for this formulation are shown in \autoref{tab:mos_results}. 
MoS models on its own increase the performance of the language model marginally.
When compared with~\autoref{tab:softmax_bottleneck}, we find that these models are worse than those that simply increases the number of embeddings.
This is expected because MoS has fewer added parameters compared to those, as it only requires several additional linear projection layers for the embeddings.

\begin{table}[t]
\centering
\small
\begin{tabular}{cccccccccHH}
\toprule
                  & $h_{ds}$ & $R$ & $\otimes$ & {}+\#params                  & PPL      & $\lambda$ & Interp. PPL & Oracle & Overlap & Helping \\ \midrule
Base LM           & -        & -   & -         & 0                            & 21.750   & -         & -           & -      & -       & -       \\
KNN               & att      & -   & L2        & $N_{ds}\times D$             & $\infty$ & 0.271     & 19.174      & 14.230 & 1.000   & 0.380   \\
KNN               & att      & -   & IP        & $N_{ds}\times D$             & $\infty$ & 0.266     & 19.095      & 14.077 & 0.890   & 0.402   \\
KNN               & ffn      & -   & L2        & $N_{ds}\times D$             & $\infty$ & 0.065     & 20.734      & 15.594 & 0.600   & 0.357   \\
KNN               & ffn      & -   & IP        & $N_{ds}\times D$             & $\infty$ & 0.050     & 21.101      & 16.254 & 0.586   & 0.340   \\
Ft. MoS+embed    & att      & 2   & IP        & $VD + 2D^2 + 2D$ & 21.986   & 0.437     & 20.720      & 17.573 & 0.522   & 0.485   \\
Ft. MoS+embed    & att      & 3   & IP        & $VD + 3D^2 + 3D$ & 22.106   & 0.422     & 20.779      & 17.609 & 0.529   & 0.474   \\
Ft. MoS Only & att      & 2   & IP        & $2D^2 + 2D$            & 22.552   & 0.371     & 21.011      & 17.796 & 0.540   & 0.465   \\
Ft. MoS Only & att      & 3   & IP        & $3D^2 + 3D$            & 22.573   & 0.371     & 21.024      & 17.812 & 0.539   & 0.464   \\
Ft. MoS Only & ffn      & 2   & IP        & $2D^2 + 2D$            & 21.351   & 0.843     & 21.338      & 20.258 & 0.553   & 0.435   \\
Ft. MoS Only & ffn      & 3   & IP        & $3D^2 + 3D$            & 21.495   & 0.733     & 21.460      & 20.322 & 0.542   & 0.472   \\
Ft. MoS Only & ffn      & 4   & IP        & $4D^2 + 4D$            & 21.321   & 0.994     & 21.321      & 20.396 & 0.569   & 0.444   \\
Ft. MoS Only & ffn      & 5   & IP        & $5D^2 + 5D$            & 21.371   & 0.909     & 21.367      & 20.406 & 0.553   & 0.439   \\ \bottomrule
\end{tabular}
\caption{Performance comparison of \knn baselines and several MoS configurations. $R$ is the number of mixtures.}
\label{tab:mos_results}
\end{table}

\subsection{Clustering Datastore}
Opposite to training the word embeddings of an increased size, we also consider ways to compress the datastore down to a similar-sized embedding matrix for softmax computation.
The intuition is that the datastore contains redundant context vectors, and thus compression could make the datastore smaller without sacrificing too much performance gain.
\cite{he2021efficient} shows that we can safely compress the datastore by clustering to 50\% of the original size without losing performance.
We test this idea further by clustering the entire datastore into a size that could fit in GPU memory (e.g. $2V$, $3V$) and thus could be easily finetuned further and use the resulting centroids to replace $W_{ds}$.
Within each cluster, there will be a distribution of different words with contexts, and we use the frequency of words within each cluster to calculate the aggregation matrix $M$ in \autoref{eqn:general}.
This would have the added benefit of ``multi-sense'' embedding, which allows similar meanings to be clustered to form a new ``meta word'' while the same word with different meanings would form different ``meta words''. 
A notable example is bank, shore, and financial institution. 
However, this does not work, mostly because of the high compression loss after clustering and the imbalanced distribution of word types among each cluster.

\section{Which Words Benefit from Approximation?}
\label{app:wordanalysis}

To further understand the unexpected results when using the different \knn approximate retrieval settings in~\autoref{sec:approximate} and~\autoref{sec:temperature}, we analyze on a token level, based on how many times each ground truth token's probability in the evaluation set are helped by each \knn setting.
It means that for each ground truth token in the evaluation, we count the times when the \knn distribution is higher than the base LM distribution $P_{LM}$, i.e., $P_{kNN} > P_{LM}$.

Since we found previously that approximate \knn provides an additional performance boost compared to ground truth \knn, we thus compare ``real mask, real score'' versus ``FAISS mask, real score'' in this analysis.
To prevent outliers, we filter out words with less than 10 occurrences in the evaluation set.
For each setting, we calculate the percentage of occurrences in the evaluation set where each token in the vocabulary  where the \knn module achieves a better probability than base LM.
We then plot the absolute difference between the percentages of the two settings, with respect to various possible attributes of the token that achieves better probability using each setting.

\begin{figure}[th]
\centering
\includegraphics[width=0.6\textwidth]{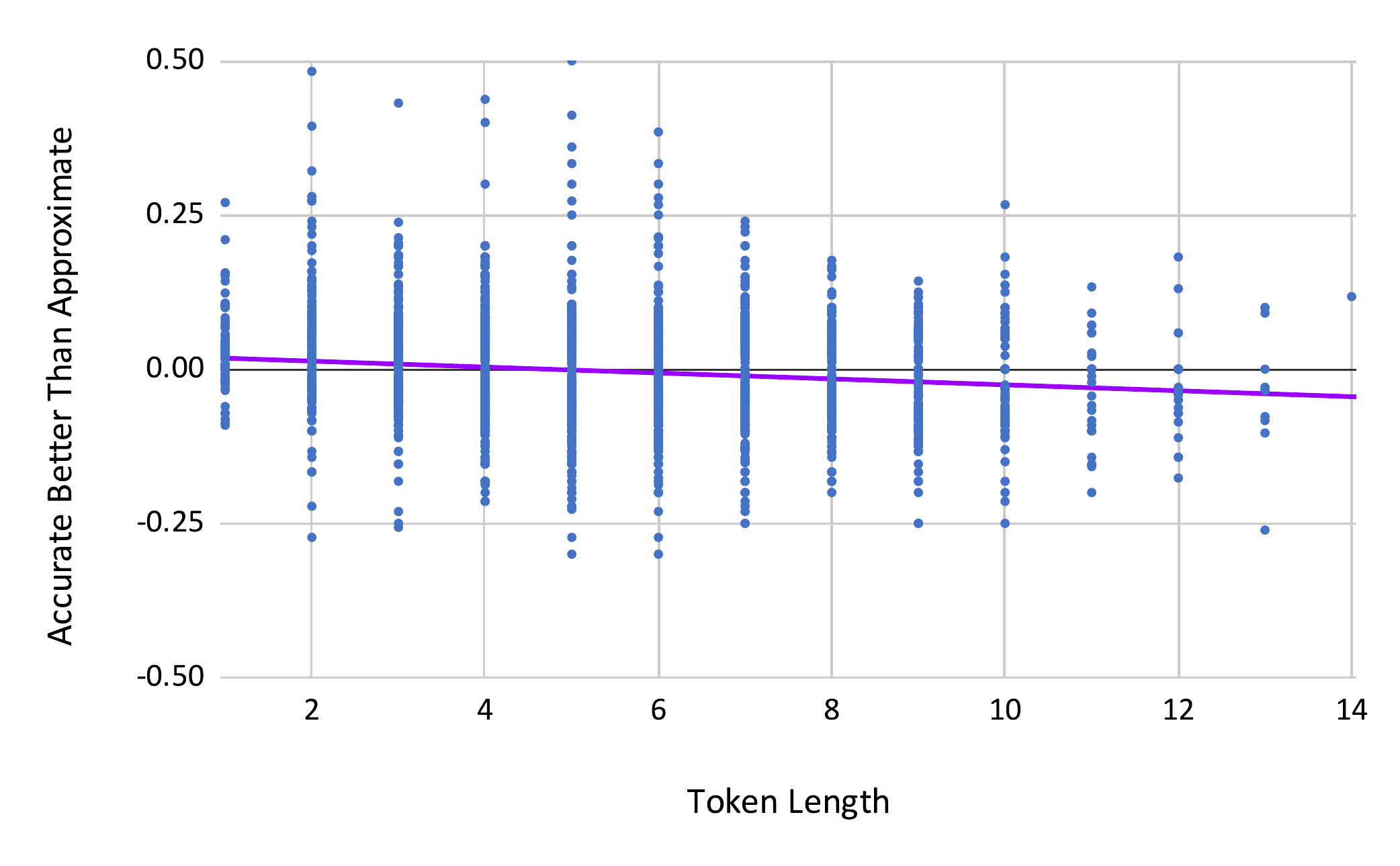}
\caption{The effect of the token character length on how much accurate nearest neighbors are better than approximate FAISS neighbors. Negative values mean worse. The trend line of the scatter points is shown.}
\label{fig:token_analysis_len}
\end{figure}

\autoref{fig:token_analysis_len} shows that the longer the token is, which usually suggests proper nouns and harder and less common words in English, are better with approximate neighbors than ground truth ones, and vice versa.
We hypothesize that this is due to longer words are more prone to overfitting in \knnlm and thus using approximate \knn provides an effect similar to smoothing and regularization.

We also compare words that could appear in more diverse contexts with words that co-occur with few distinct contexts.
To measure how diverse the contexts of each word in the vocabulary is, we calculate both the forward and backward bigram entropy for each word in the evaluation set that has more than 10 occurrences.
The bigram entropy is a simple yet good indicator of context diversity for a given word, as used in Kneser–Ney smoothing~\citep{ney1994structuring}.
We calculate both the forward and backward bigram entropy for each word $w$ as follows, where $w_\text{after}$ and $w_\text{before}$ represent the word after and before the given word $w$.
\begin{align}
    H_\text{forward}(w) & = -\sum_{w_\text{after}} p(w_\text{after}|w) \log p(w_\text{after}|w) \\
    H_\text{backward}(w) & = -\sum_{w_\text{before}} p(w_\text{before}|w) \log p(w_\text{before}|w)
\end{align}
Forward and backward entropy represents how diverse the context after and before the given word is.
Intuitively, bigram entropy is supposed to indicate words that can appear in lots of different contexts.
The higher the entropy of a word, the more diverse its context is, and vice versa.
For example, words like ``Francisco'' would have a low entropy because it mostly comes after ``San''.

\begin{figure}[th]
\centering
\includegraphics[width=0.49\textwidth]{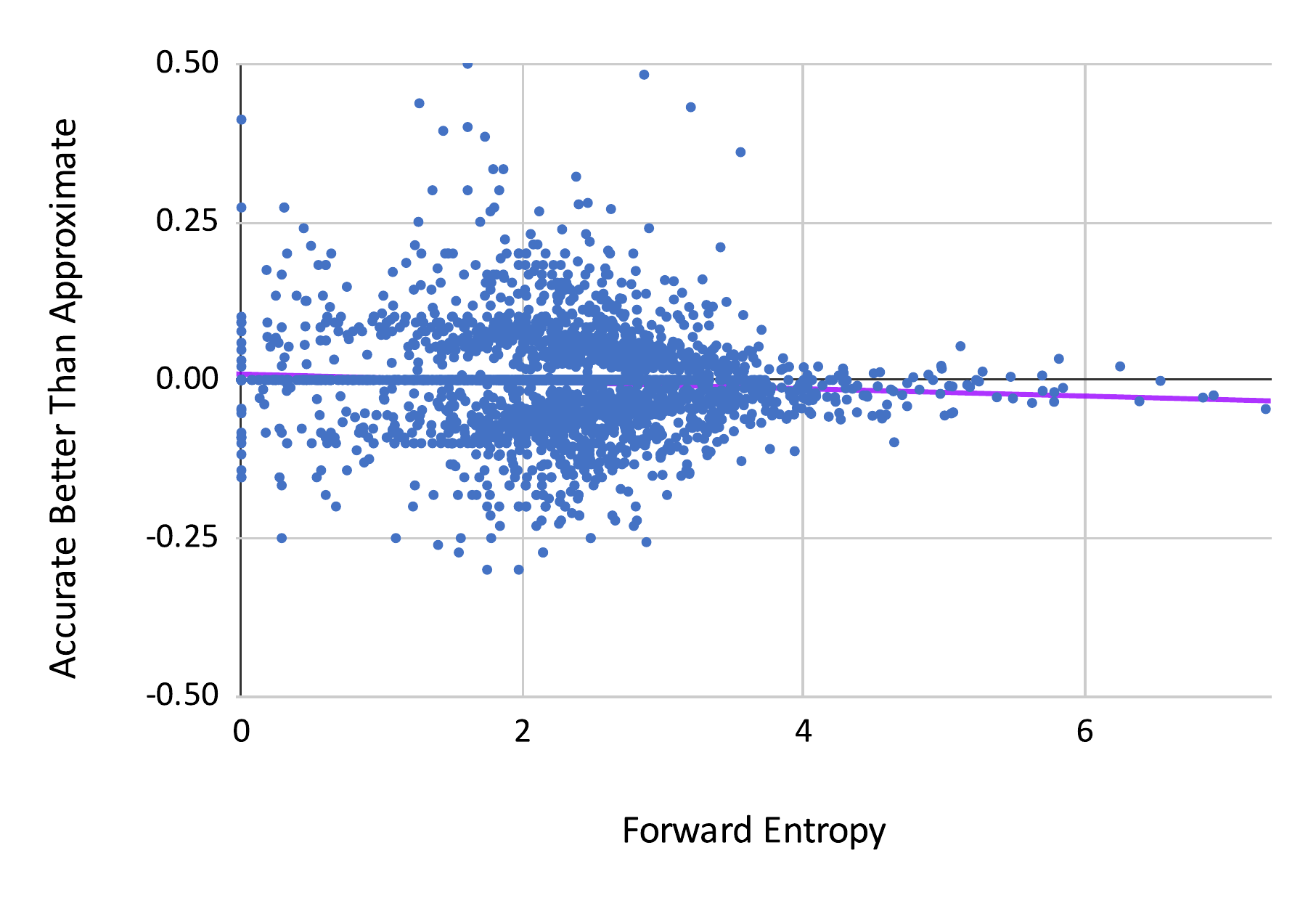}
\includegraphics[width=0.49\textwidth]{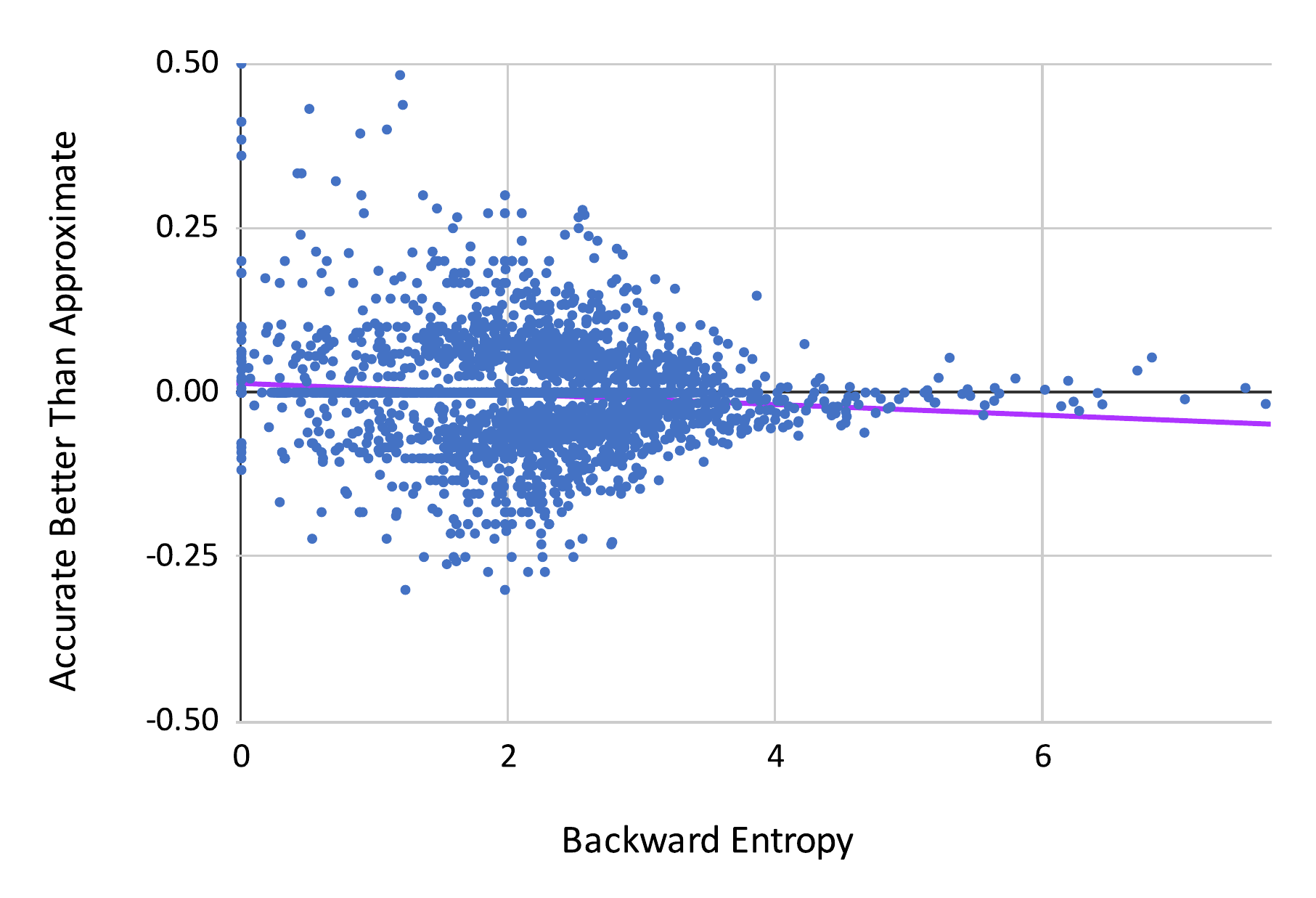}
\caption{The effect of the forward and backward entropy of words on how accurate nearest neighbors are better than approximate FAISS neighbors. Negative values mean worse. The trend line of the scatter points are shown.}
\label{fig:entropy}
\end{figure}

The comparison is shown in~\autoref{fig:entropy}.
We can see that the higher the entropy in both forward and backward cases, the better using approximate nearest neighbor search becomes.
This suggests that words that appear in many different contexts are better off with an approximate \knn, and ``easy-to-predict'' examples such as ``Jersey'' and ``Fransisco'' is  better with accurate \knn, possibly because these examples are less prone to overfitting errors and thus requires less regularization from approximation.

\section{Failed Hypotheses}
\label{app:failed}
\subsection{Distance Metric}
\label{app:dist_metric}
We hypothesize that the key to \knnlm's performance gain is the ensemble of two distance metrics: the standard dot product distance (which the LM uses) with the L2 distance (which the \knn component uses as $\otimes$).
We tried to replace the \knn component with a component that just takes the tokens retrieved by the \knn search and returns their L2 distance to the LM output word embeddings: $W_{sm}\otimes h_{ds}$ instead of $W_{ds}\otimes h_{ds}$, where $\otimes$ represents the negative L2 distance.
We tried this with both variants of $h_{ds}$, attention layer output, and feedforward layer output. None of these helped. 

\subsection{Sparsification}
\label{app:sparsification}
In \autoref{eqn:general}, $\text{mask-to-k}(\cdot)$ used by \knn retrieval induces sparsity in the distribution over the vocabulary, due to a small $k$ compared to the number of vocabulary $V$.
We hypothesize that the in \knnlm, the \knn distribution is sparse, practically increasing the probability of the top-$k$ entries.
The \knn distribution has up to 1024 entries that are non-zero, concentrating more probability mass over the most likely tokens.
This effect is similar to the redistribution of probability mass for text generation in~\cite{holtzman2019curious}.
We test this hypothesis only by taking top 32, 64, 128, 512, or 1024 tokens in the parametric LM probability and zeroing out the probabilities of the rest of the tokens. 
To compensate, we experiment with different softmax temperatures and then interpolate with the parametric LM probability.
This isolates the effect of the datastore and retrieval at all, and this does not help at all, suggesting that sparsification of the output probability alone is not enough.

Another attempt is to hypothesize that the key  in \knnlm is that it selects ``which tokens to include'' in the \knn distribution, and not their distances.
The intuition behind is that maybe the selection of the top tokens according to the \knn search is better than that from the dot-product distance between the language model's output vector and all the vocabulary embeddings.
We perform experiments similar to the previous attempt, sparsifying the output probability with the tokens retrieved by the \knn search (but ignoring the distances provided by the \knn search) rather than the top $k$ tokens of the LM, with and without removing duplicates.  
In the best case, they manage to reduce the perplexity by 0.5 (whereas \knnlm reduces by nearly 2).

\subsection{Location within Context Window}
Supposedly, words in the beginning of the ``context window'' of the transformer at test time have less contextual information than words toward the end of context window.

We hypothesized that maybe the base LM performs worse in one of these (beginning vs. end of the context window), and maybe \knnlm provides a higher improvement in one of these. 
We measured the per-token test perplexity with respect to the location of each token in the context window.
However, we did not find any significant correlation between the performance of the base LM and the location, and no significant correlation between the difference between \knnlm and the base LM and the location.

We also hypothesized that maybe the beginning of every Wikipedia article is more ``predictable'', and the text becomes more difficult to predict as the article goes into details.
However, we also did not find any correlation with the location of the word within the  \emph{document} it appears in.

\subsection{Stolen Probabilities}
\label{app:stolen_prob}
The \emph{stolen probabilities} effect \citep{demeter2020stolen} refers to the situation where the output embeddings of an LM are learned such that some words are geometrically placed \emph{inside} the convex hull that is formed by other word embeddings.
Since language models generate a score for every output word by computing the dot product of a hidden state with all word embeddings, \citet{demeter2020stolen} prove that in such a case, it is impossible for words inside the convex hull to be predicted as the LM's most probable word (the ``argmax''). 

We hypothesized that \knnlm solves the stolen probabilities problem by allowing to assign the highest probability to \emph{any} word, given a test hidden state that is close enough to that word's datastore key. 
Nevertheless, as shown by \citet{grivas2022low}, although this problem might happen in small RNN-based language models, in modern transformers it rarely happens in practice.
Using the code of \citet{grivas2022low}, we checked the embeddings matrix of our model and of the checkpoint provided by \citet{khandelwal20generalization}. Indeed, we found that in both models -- \emph{no word is un-argmaxable}.

\subsection{Are \knnlm Just Ensembling?}
\label{app:ensembling}

Our hypothesis is that \knn component only provides another model for ensembling.
The interpolation process is basically an ensemble model.
Technically it is unsurprising that \knnlm will have the benefit from ensembling, but we perform experiments to see how it compares to other ensembling.
We trained another language model with the same architecture as the base LM we used throughout the experiments, with some variants having more than one embedding vector for each word (similar to \autoref{sec:bottleneck}).
We interpolate the models with the original base LM, and the results are shown in \autoref{tab:ensembling}.
We can see that even just ensembling the base LM with another identical model, but trained with a different random seed, provides a huge performance boost, both on interpreted perplexity and on oracle perplexity.

\begin{table}[h]
\centering
\small
\begin{tabular}{ccccccHccHH}
\toprule
Prev. Layers & $h_{ds}$ & $N_{ds}$ & $\otimes$ & {}+\#params        & PPL      & $\lambda$ & Interp. & Oracle & Overlap & Helping \\ \midrule
same         & -        & -        & -         & 0                  & 21.750   & -         & -           & -      & -       & -       \\
same         & att      & Big      & L2        & $N_{ds}\times D$   & $\infty$ & 0.271     & 19.174      & 14.230 & 1.000   & 0.380   \\
same         & att      & Big      & IP        & $N_{ds}\times D$   & $\infty$ & 0.266     & 19.095      & 14.077 & 0.890   & 0.402   \\
same         & ffn      & Big      & L2        & $N_{ds}\times D$   & $\infty$ & 0.065     & 20.734      & 15.594 & 0.600   & 0.357   \\
same         & ffn      & Big      & IP        & $N_{ds}\times D$   & $\infty$ & 0.050     & 21.101      & 16.254 & 0.586   & 0.340   \\
diff         & ffn      & 1x       & IP        & $F$ + $V\times D$  & 21.569   & 0.497     & 18.941      & 14.980 & 0.592   & 0.474   \\
diff         & ffn      & 2x       & IP        & $F$ + $2V\times D$ & 21.914   & 0.467     & 18.948      & 14.885 & 0.602   & 0.444   \\
diff         & ffn      & 3x       & IP        & $F$ + $3V\times D$ & 22.206   & 0.457     & 18.981      & 14.853 & 0.601   & 0.463   \\ \bottomrule
\end{tabular}
\caption{Performance comparison of \knn baselines and models with different size output embeddings re-trained from scratch.}
\label{tab:ensembling}
\end{table}

\begin{figure}[h]
\centering
\includegraphics[width=0.6\textwidth]{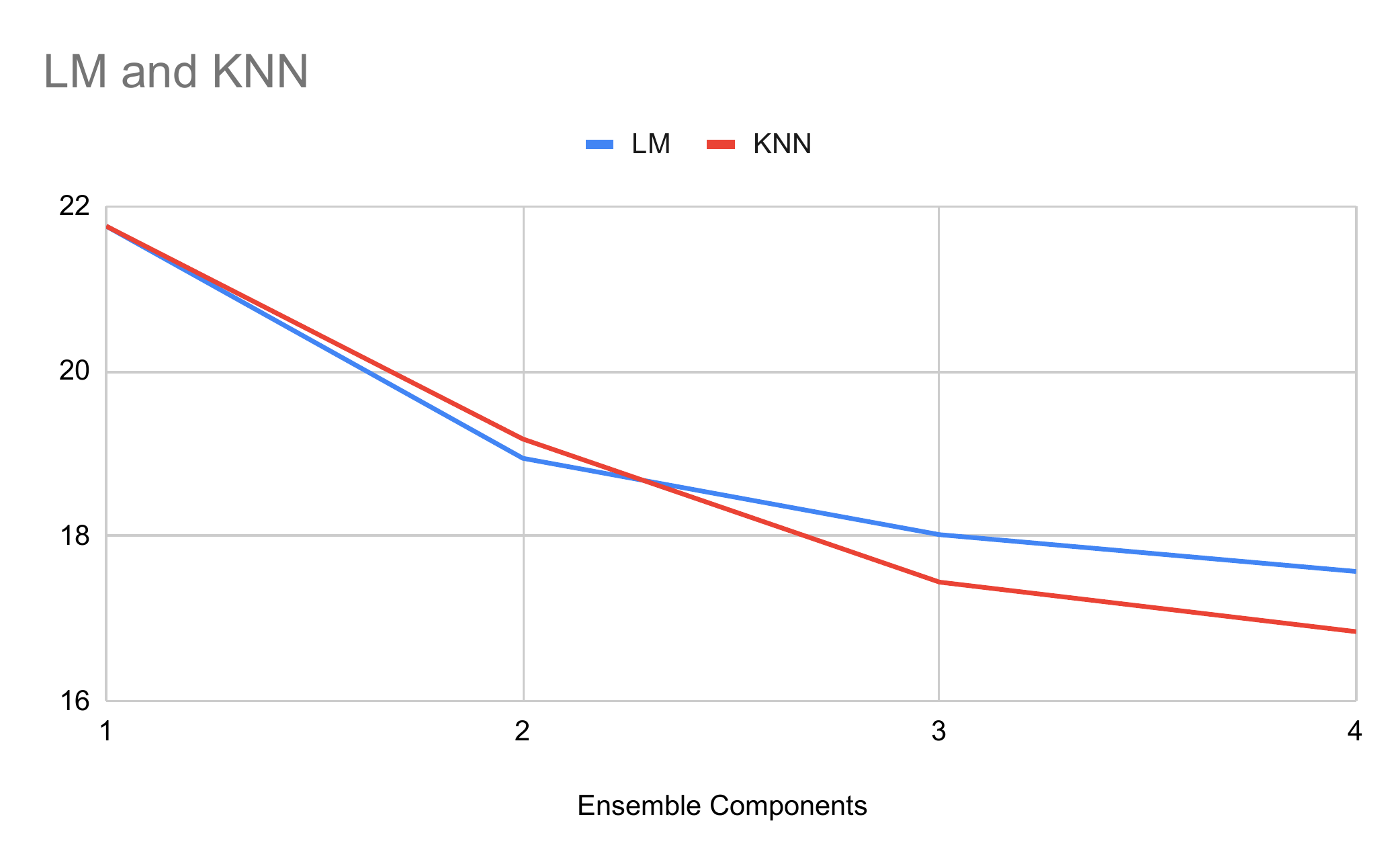}
\caption{Ensembling effect comparison, between multiple base LMs and multiple base LMs plus \knn component.}
\label{fig:ensemble}
\end{figure}

However, just because ensembling two LMs of the same architecture provides better performance than interpolating the base LM with \knn does not necessarily suggest that  \knn's performance improvement can be fully replaced by model ensembling.
In other words, we are interested in whether the  \knn performance improvements are orthogonal to that of model ensembling.
To test this, we compare the performance of the ensemble of $K$ multiple LMs versus the ensemble of $K-1$ multiple LMs plus the  \knn component.
The comparison is fair because we have the same number of models in the ensemble, and the only difference is whether the  \knn component is included.
The results are shown in \autoref{fig:ensemble}.
For the ``LM'' series, each point is $K$ LMs ensemble, and for the ``\knn'' series, each point is $K-1$ LMs plus \knn.
We can see that even at 4-ensemble, the ensemble that contain \knn as a component still have a considerable edge over the 4-ensemble that contain just LMs.

\subsection{Are \knnlm Just Alternative Training Methods?}

\subsubsection{Overfitting}
\label{app:overfitting}

\begin{table}[h]
\centering
\small
\begin{tabular}{cccccccHccH}
\toprule
                   & Prev. Layers & $h_{ds}$ & $N_{ds}$ & $\otimes$ & {}+\#params       & PPL      & $\lambda$ & Interp. & Oracle & Helping \\ \midrule
Base LM            & same         & -        & -        & -         & 0                 & 21.750   & -         & -           & -      & -             \\
KNN                & same         & att      & Big      & L2        & $N_{ds}\times D$  & $\infty$ & 0.271     & 19.174      & 14.230 & 0.380         \\
KNN                & same         & att      & Big      & IP        & $N_{ds}\times D$  & $\infty$ & 0.266     & 19.095      & 14.077 & 0.402         \\
KNN                & same         & ffn      & Big      & L2        & $N_{ds}\times D$  & $\infty$ & 0.065     & 20.734      & 15.594 & 0.357         \\
KNN                & same         & ffn      & Big      & IP        & $N_{ds}\times D$  & $\infty$ & 0.050     & 21.101      & 16.254 & 0.340         \\
Overfit@92  & diff         & ffn      & $V$      & IP        & $F$ + $V\times D$ & 1702.806 & 0.010     & 21.732      & 17.764 & 0.362         \\
Overfit@129 & diff         & ffn      & $V$      & IP        & $F$ + $V\times D$ & 8966.508 & 0.010     & 21.733      & 17.814 & 0.349         \\ \bottomrule
\end{tabular}

\caption{Performance comparison of several baselines with two overfitted models, at 92 and 129 additional epochs.}
\label{tab:overfit}
\end{table}

Since \knnlm improves perplexity even with the same training dataset as datastore, we are curious if \knnlm works by only ``memorizing'' the training data.
The hypothesis is that the datastore and the \knn search are trying to memorize the training data. 
In other words, the parametric LM is under-fitting some tokens.
The intuition behind this is that the \knn component retrieves examples directly from the training set. 
What if we could retrieve the same examples using an overfitted LM?
We took the trained LM, removed the dropout, and continued training until almost perfect fit (very small training loss).
We then interpolated the overfitted transformer with the original LM. 
The results are shown in \autoref{tab:overfit}.
$F$ represents the number of parameters in the base LM, minus the output embedding matrix.
We can see that overfitting can provide very little help after interpolation.
Looking at the oracle performance, we think that the overfitted model memorizes some rare contexts and tokens in the training set where it could be useful during evaluation.
However, the overfitting hurts the performance on other tokens too much so that even interpolation is not able to balance the performance.

\subsubsection{Soft-Label Training}
\label{app:soft_label}
\cite{yang2022nearest} claims that using ``soft labels'' during training is the key to \knn's success, that interpolates the ground truth labels with \knnlm model outputs, effectively ``distilling'' \knnlm.
It is based on the hypothesis that the room for \knnlm's improvement over base LM lies in the ``over-correction'' when training with a 1-hot labels.
This is related to the effect from label smoothing methods~\citep{szegedy2016rethinking,pereyra2017regularizing,meister2020generalized}.
However, we believe that this explanation is not satisfactory. 
If the key is training with soft-labels, why do these soft labels must be provided specifically by a \knn search?
If soft labels were the key, then soft-label training where the labels come from the base LM itself should have worked as well.
To separate the effect of soft labeling from the \knn's additional guidance, we train another LM with the same model architecture as the base LM, with the soft labels from the base LM.
This teacher-student training is to distill the knowledge from the base LM~\citep{hinton2015distilling}. 
We find that by just training with ``soft labels`` from the base LM to alleviate the alleged ``over-correction'' problem is not the key, as this does not help with the interpolated perplexity at all.
This suggests that even with the same training data, \knn still provides valuable additional guidance.

\subsubsection{Training to Optimize Interpolated Loss}
\label{app:interpolated_loss}
In \autoref{sec:bottleneck}, we discover that using over-parameterization with standard LM training loss does not further close the gap towards \knnlm. This suggests that some regularization term may be needed during training to make the multiple embeddings not converge to the same vector, rendering over-parameterization useless.

From~\autoref{tab:input_representation}, we see that a better interpolated perplexity may not require a very low perplexity when measured only with the extra input representation.
However, we still use a standard LM loss to only train the additional embedding matrix, that directly minimizes the perplexity using only the extra input representation.
This discrepancy between training and the evaluation with interpolation suggests that training with an alternative loss function that interpolates the base LM's output with the output using the extra input representation may be beneficial.

To test the hypothesis that standard LM training loss do not emphasize the examples where base LM performs badly, we train the extra model's parameter $W_{ds}$, with interpolated loss $L$: 
\begin{equation}
    L =\text{CrossEntropy}(\lambda\text{softmax}(W_{ds} \cdot h_{ds}) + (1-\lambda)\text{softmax}(W_{sm} \cdot h_{sm}), y)
\end{equation}
$y$ represents the ground truth label for each context.
We only learn the parameter $W_{ds}$ while freezing all other parameters, similar to all other experiments.
We choose $\lambda=0.25$ as it is the best hyper-parameter for \knnlm experiments and our goal for this training is to mimic the loss of \knnlm after interpolation.
This training loss effectively assigns a higher value to the training examples where the base LM's loss is high, suggesting the need for the extra $W_{ds}$ to help with these hard cases.
However, for either ``att'' for ``ffn'' for $h_{ds}$, either $V$ or $3V$ for the number of embeddings in $W_{ds}$, we are unable to achieve a better perplexity than just the base LM.
This suggests that, while nice on paper, the interpolated loss optimization process is not trivial.

\end{document}